\documentclass[journal]{IEEEtran}

\usepackage{times}
\usepackage{epsfig}
\usepackage{graphicx}
\usepackage{amsmath}
\usepackage{amssymb}
\DeclareMathAlphabet\mathbfcal{OMS}{cmsy}{b}{n}

\usepackage{listings}

\usepackage{xcolor}

\definecolor{codegreen}{rgb}{0,0.6,0}
\definecolor{codegray}{rgb}{0.5,0.5,0.5}
\definecolor{codepurple}{rgb}{0.58,0,0.82}
\definecolor{backcolour}{rgb}{0.95,0.95,0.92}

\lstdefinestyle{mystyle}{
    backgroundcolor=\color{backcolour},   
    commentstyle=\color{codegreen},
    keywordstyle=\color{magenta},
    numberstyle=\tiny\color{codegray},
    stringstyle=\color{codepurple},
    basicstyle=\ttfamily\footnotesize,
    breakatwhitespace=false,         
    breaklines=true,                 
    captionpos=b,                    
    keepspaces=true,                 
    numbers=left,                    
    numbersep=5pt,                  
    showspaces=false,                
    showstringspaces=false,
    showtabs=false,                  
    tabsize=2
}

\usepackage{mathtools}

\DeclarePairedDelimiter\floor{\lfloor}{\rfloor}

\usepackage[utf8]{inputenc}
\usepackage[english]{babel}
\usepackage[inline]{enumitem}
\usepackage{subcaption}
\usepackage{hyperref}
\usepackage{diagbox}

\ifCLASSINFOpdf

\else

\fi

\begin{document}

\title{Context-self contrastive pre-training for crop type semantic segmentation}

\author{Michail Tarasiou, Riza Alp G\"{u}ler,
        and~Stefanos Zafeiriou}% <-this % stops a space

\markboth{Journal of \LaTeX\ Class Files,~Vol.~14, No.~8, August~2015}%
{Shell \MakeLowercase{\textit{et al.}}: Bare Demo of IEEEtran.cls for IEEE Journals}

\maketitle

\begin{abstract}
In this paper, we propose a fully supervised pre-training scheme based on contrastive learning particularly tailored to dense classification tasks. The proposed {\it Context-Self Contrastive Loss} (CSCL) learns an embedding space that makes semantic boundaries pop-up by use of a similarity metric between every  location in a training sample and its local context.
For crop type semantic segmentation from {\it Satellite Image Time Series} (SITS) we find performance at parcel boundaries to be a critical bottleneck and explain how CSCL tackles the underlying cause of that problem, improving the state-of-the-art performance in this task. Additionally, using images from the {\it Sentinel-2} (S2) satellite missions we compile the largest, to our knowledge, SITS dataset densely annotated by crop type and parcel identities, which we make publicly available together with the data generation pipeline. Using that data we find CSCL, even with minimal pre-training, to improve all respective baselines and present a process for semantic segmentation at greater resolution than that of the input images for obtaining crop classes at a more granular level. The code and instructions to download the data can be found in {\color{magenta} \verb{https://github.com/michaeltrs/DeepSatModels{\color{black}.}

\end{abstract}

\begin{IEEEkeywords}
Deep Learning, Pre-training, Contrastive Learning, Convolutional Neural Networks, Self-Attention, Sentinel-2, Semantic Segmentation, Crop Type Segmentation.
\end{IEEEkeywords}

\IEEEpeerreviewmaketitle

\begin{figure}[!h]
\begin{center}
   \includegraphics[width=0.5\textwidth]{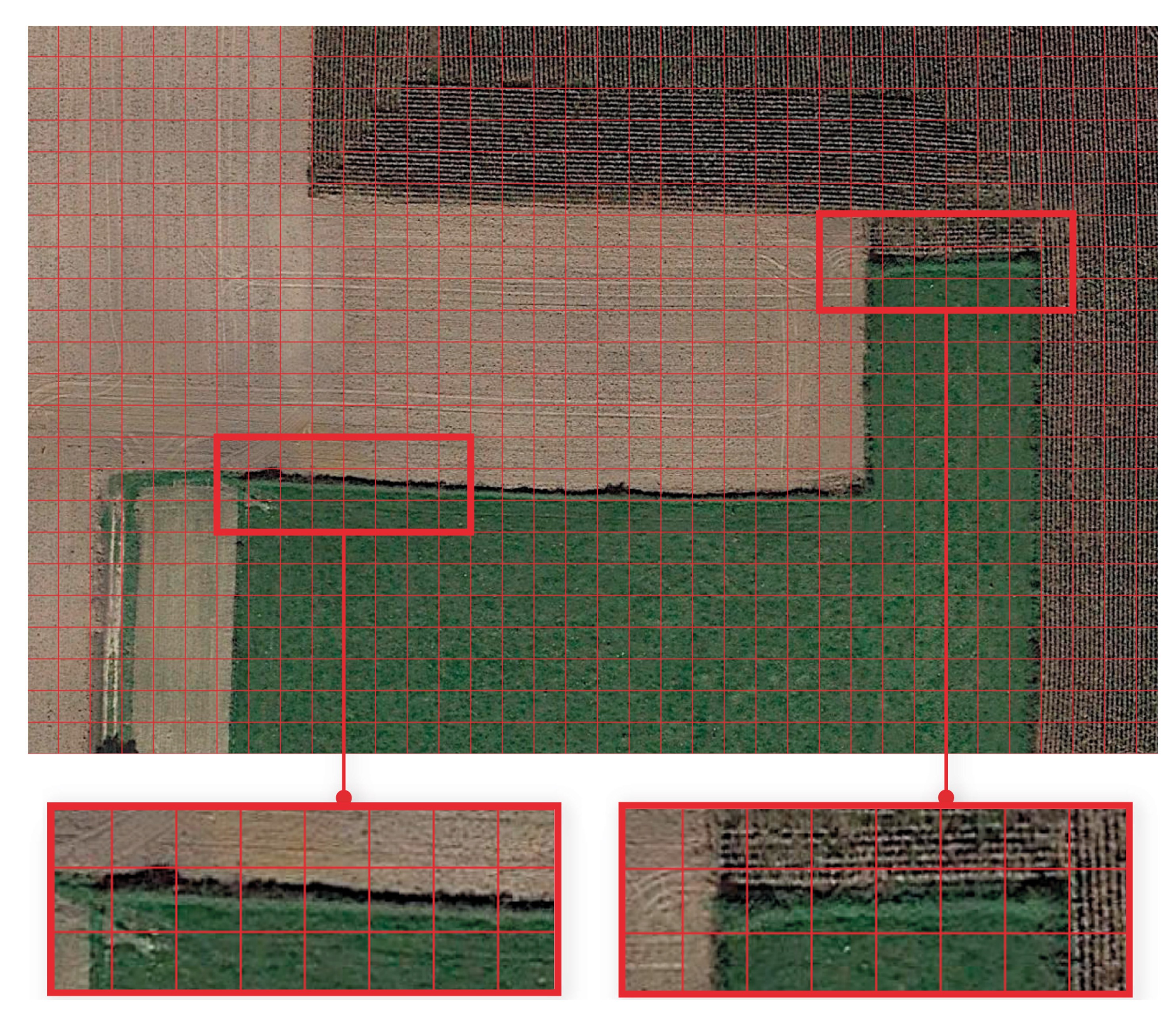}
\end{center}
    \caption{Pixel size is coarse enough for the signal at boundaries to mix. Here the 10m resolution {\it Sentinel-2} grid is overlaid on a high-resolution image from \href{https://www.google.com/maps/place/46\%C2\%B026'24.5\%22N+5\%C2\%B000'57.8\%22E/@46.439588,5.0160048,714m/data=!3m1!1e3!4m5!3m4!1s0x0:0x0!8m2!3d46.4401361!4d5.0160528}{Google Earth}. The magnified regions show grid locations that contain signal from multiple crop types and objects. The proposed {\it Context-Self Contrastive Loss} directly compares computed embeddings in local neighbourhoods. As such, interior regions naturally have a small contribution to the overall loss while at the same time the network learns to disambiguate boundary pixels by comparing them to nearby locations.}
    \label{teaser1}
\end{figure}

\begin{figure*}[!h]
\begin{center}
   \includegraphics[width=0.975\textwidth,  trim={0 12cm 0 6cm}]{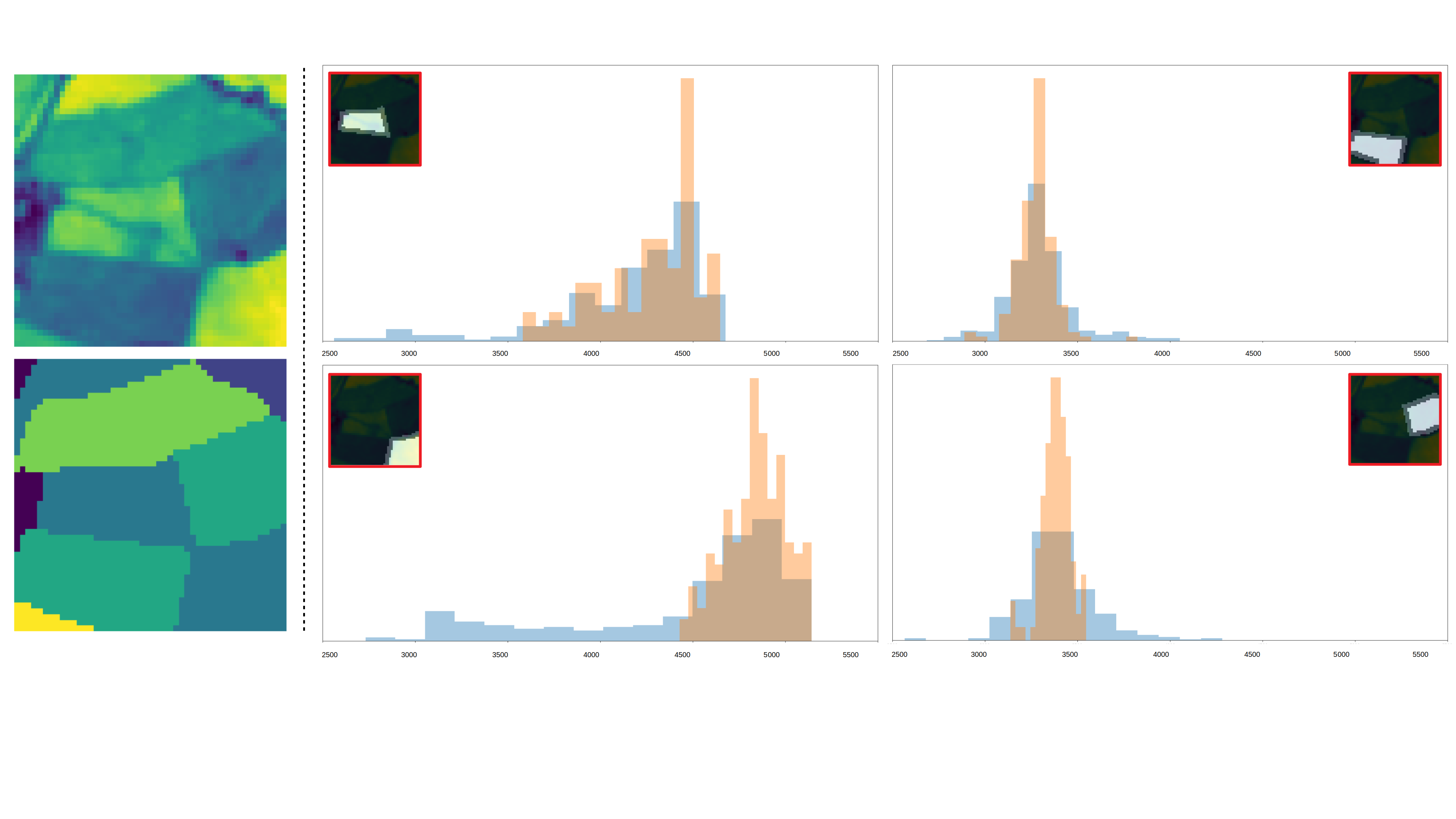}
\end{center}
    \caption{In satellite images interior regions of agricultural fields consist of large homogeneous regions with little signal variation. (top-left) Spectral band B08 for a location in France, (bottom-left) ground truth crop type maps for the same location. (right) Histograms of intensities over whole parcel regions (blue) and interior points (orange). Each histogram presents the intensities for a specific field as indicated by the respective mask (red outline). Interior, interior+boundary and exterior locations are shown in white, grey and black colors respectively. We observe that most of the variability in intensities originates from the boundary regions. The distribution of intensities for other spectral bands is similar to that of band B08.}
    \label{teaser2}
\end{figure*}

\section{Introduction}
The large availability of imagery made available by {\it Earth Observation} (EO) satellites has enabled the development of tools for the automated monitoring of food supplies and the design and control of policies aiming at agricultural development. In the {\it European Union} (EU) agricultural subsidies provided as part of the {\it Common Agricultural Policy} (CAP) comprised approximately 39$\%$ of last year's overall budget at \texteuro 58 billion \cite{cap}. Distribution of these subsidies to over 10 million farms is currently performed by use of crop maps which are compiled by parcel topographies and crop types declared by EU farmers. New reforms of the CAP are expected to place particular emphasis on the use of automated tools for EO. In particular, the {\it Sentinels for Common Agricultural Policy} (Sen4CAP) project aims at making use of the {\it European Space Agency} (ESA) {\it Sentinel} satellites for providing stakeholders with validated algorithms, products and tools relevant for the management of the CAP. Provided examples for CAP monitoring of agricultural practices \footnote{http://esa-sen4cap.org/content/agricultural-practices} and crop diversification requirements \footnote{http://esa-sen4cap.org/content/crop-diversification} both impose thresholds on the area of declared arable land dedicated to different crop types. Their accuracy should benefit greatly from the development of automated crop identification processes at high granularity which is the goal of this work.

However, a performance analysis of state-of-the-art large-scale crop-type identification systems presented in section \ref{boundary_perf} showed us that the classification accuracy varies with the location within a parcel and drops significantly near the parcel boundaries. This is in part expected for all dense labelling tasks and particularly so for the segmentation of S2 images in which location can be a few pixels off with respect to the ground truths, introducing label noise at object boundaries. We identify an additional difficulty in correctly classifying parcel boundaries to the fact that satellites dedicated to monitoring land surface variability trade off resolution with high revisit time. As a result pixel size is coarse enough for the signal from different crop types or objects to mix. This is demonstrated in Fig.\ref{teaser1}, where we overlay the resolution at hand on high resolution images. The highest resolution band of S2 images covers a 10m$\times$10m region per pixel. It is worth pointing out that this pixel-size is large enough to include complete objects such as roads, which adds further complexity to the problem. Additionally, the temporal patterns of plant growth during a growth year can offer significant cues for crop type identification. Models employed for this task have been shown to benefit from spending their computational budget on capturing the temporal dimension of the input data \cite{spacetime}. However, as discussed above, the use of SITS poses a limit on spatial resolution, making medium resolution images captured at a high temporal frequency an indispensable ingredient for crop type identification. These could be combined with high resolution images at low temporal granularity which would not be as informative in terms of crop identification but would contain less signal mixing at the object boundaries. Despite that, we argue that there is value in techniques that enhance performance of models using coarse resolution images alone for two reasons: 1) these are simpler models that do not require fusion techniques to incorporate heterogeneous data, 2) such techniques are likely to also benefit models that make use of both high and low resolution data. 
We also note that interior locations of agricultural parcels in satellite images include large homogeneous regions with little structure and small signal variation as shown in Fig.\ref{teaser2}. Given this observation the case could be made in favour of a two stage detect-then-segment method which detects agricultural parcels, identifies the crop type based on a set of near-center pixels and then segments the extent of the parcel using a class-agnostic segmentation head. However, there are two main issues: 1) the problem of correctly classifying boundaries still remains, a two stage approach would simply shift the onus of correctly classifying boundaries from the classifier to the class-agnostic segmentation head, 2) choosing to detect, classify and segment parcel instances using a single model would dedicate part of that model's capacity in identifying parcel instances which could lead to lower performance if the size of the model is not increased.

For crop classification at the parcel level \cite{garnot2019satellite} argue that S2 pixel size is coarser than the typical agricultural textural information and show that using spatial modelling is not at all critical for good performance. What would be the best strategy though if we are interested in predicting crop types for every pixel and moreover for boundary pixels which are the most severely affected by the coarse size? 
We argue that class attribution near boundaries can be improved by learning to detect semantic boundaries by contrasting each pixel with their neighbourhood pixels. As discussed, interior locations of agricultural parcels include large homogeneous regions with little structure and small signal variation. While the {\it Cross-Entropy} (CE) criterion still needs to assign a class to all interior pixels a contrastive criterion constrained to local neighbourhoods naturally contains a small loss contribution from these regions as they are encoded by similar embeddings. The above two observations constitute the motivation behind our proposed method.
Our key contributions are the following: \begin{enumerate}
    \item Inspired by recent developments on visual self-attention we design a contrastive learning pre-train scheme, targeting performance at the parcel boundaries (section \ref{proposed_method}). Applying the proposed approach on a publicly available crop segmentation dataset, we raise the state-of-the-art {\it mIoU} from $78.75\%$ to $81.03\%$ (section \ref{compare_sota}). 
    \item Regional variations of plant growth patterns is a major challenge hindering generalization. We present the largest, measured by {\it Area of Interest} (AOI), {\it Time Period of Interest} (POI) and number of crop types, dataset for crop type segmentation which we make publicly available together with the data generation pipeline to facilitate future research (section \ref{dataset}). We benchmark this dataset with state-of-the-art models which we consistently outperform using our pre-training scheme (section \ref{compare_sota}).
    \item We present a simple method for segmenting crops from satellite images at a higher resolution than the input by leveraging available ground truths and show performance improvements in segmenting crops at $\times 4$ the input resolution over training at base resolution (section \ref{super_res}).

\end{enumerate}

\section{Background}\label{sec:background}

{\bf Notation: }we denote {\it scalars} as lower-case letters $x$, {\it vectors} as bold lower case letters $\mathbf{x}$, {\it matrices} as bold upper-case letters $\mathbf{X}$, and {\it higher order tensors} as bold calligraphic letters $\mathbfcal{X}$. The number of dimensions of a tensor is the order of that tensor. An element $(i,j)$ of an order 3 tensor $\mathbfcal{X} \in \mathbb{R}^{I_1 \times I_2 \times I_3}$ is indicated as a vector $\mathbf{x}_{ij}$. An element $(i,j)$ of an order-4 tensor $\mathbfcal{X} \in \mathbb{R}^{I_1 \times I_2 \times I_3 \times I_4}$ is indicated as a matrix $\mathbf{X}_{ij}$. The tensor-dot product ($\bullet$) of two tensors is defined as a contraction with respect to the last index of the first one and the first index of the second one. For example, the tensor-dot product of an order-3 tensor $\mathbfcal{W} \in \mathbb{R}^{I_1 \times I_2 \times d}$ and a vector $\mathbf{x} \in \mathbb{R}^d$ is matrix $\mathbf{Y}=\mathbfcal{W} \bullet \mathbf{x} \in \mathbb{R}^{I_1 \times I_2}$ with elements $y_{ij} = \sum_{k=1}^d \mathbf{w}_{ijk} \cdot x_k$.\\

\textbf{Crop type identification from satellite images} is a challenging task that involves assigning one of $C$ crop categories at a set of desired locations on a geospatial grid, most commonly at the agricultural parcel or pixel level.
Rather than  using single images as inputs crop identification has been shown to work best by modelling the temporal patterns of plant growth during a growing season \cite{spacetime}.
{\it Remote Sensing} approaches to crop type identification have included multiple preprocessing steps, adding expert knowledge to correct satellite image capturing and extract hand-crafted features, e.g. various vegetation indices \cite{dvi} used as input to machine learning classifiers \cite{cc1, ndvi1, ndvi2, ndvi3, hmm, pel_rand}. More recently {\it Deep neural Networks} (DNN) have achieved increasingly competitive performance outperforming previous approaches by a large margin \cite{Ruwurm1,Ru_wurm_2018,dnn2, dnn3, dnn4}. These methods make use of minimal data preprocessing and are shown to handle challenges such as cloudy images, image noise and atmospheric corrections automatically during network feature extraction \cite{Ru_wurm_2018, Rustowicz2019SemanticSO, Ruwurm2}. Some works involve temporal modelling of single pixel or parcel level aggregated features \cite{Ruwurm1, pel1, garnot2019satellite} while others jointly capture temporal and spatial patterns \cite{Ru_wurm_2018, duplo, Rustowicz2019SemanticSO, spacetime}. Fewer works have tackled semantic segmentation of crop types \cite{Ru_wurm_2018, Rustowicz2019SemanticSO, garnot_iccv}.

In particular \cite{Rustowicz2019SemanticSO} propose two types of models: a fully convolutional {\it 3D-UNET} \cite{3dunet} and a {\it 2D-UNET} feature extractor coupled with a {\it CLSTM} decoder \cite{conv_lstm} both achieving state-of-the-art performance. In our work we use a novel contrastive pre-training scheme to significantly improve these results, particularly so for locations at parcel boundaries. 

On super-resolution with satellite images \cite{malkin} present a method for semantic segmentation at the resolution of input images utilizing only low-resolution ground-truth labels and assuming a known joint distribution between low and high-resolution ground truths. 
Our method utilizes vector representations of parcel geometries, a common output of agricultural field studies, making no further assumptions. It is also different in that we predict classes at a higher resolution than inputs.

\textbf{Contrastive learning} aims at mapping data points to an embedding space in which a distance metric encodes the similarity between inputs, with similar samples brought closer together and dissimilar samples moved far apart. It can be viewed as a subcategory of metric learning in which the loss function operates directly on the embedding space and is driven by known (supervised setting) or assumed (self-supervised setting) similarities between samples \cite{metric_learn}. A classification loss, e.g. CE loss, can also be viewed as a metric style loss as it constructs an embedding space which preserves class similarity as distance. However, it does so indirectly by transforming the embedding vector into logits through matrix multiplication with a learnt weight matrix and maximizing the likelihood that the sample belongs in the ground truth class. In comparison, a contrastive loss compares the embeddings produced by different samples and aims at representing sample similarity directly as distance in embedding space.

The seminal work of \cite{chopra05} for the face verification task used a {\it Siamese Neural Network} \cite{siamese} architecture to process two input samples with known semantic similarity and applied a contrastive loss function directly on extracted features. Similarly, a {\it Triplet Loss} \cite{triplet2,triplet} uses three samples at a time, an anchor, a positive and a negative sample and aims at making the anchor-positive distances smaller and the anchor-negative distances larger.  

Several works have employed contrastive learning to enhance the discriminatory capacity of neural networks in downstream classification tasks. This can be achieved by either learning a linear classifier directly on the latent space \cite{instance_discr, cpc, contr_loss_chen, momentum_contrast, momentum_contrast2, densepretrain1}, or by using network weights as a good initialization point for further end-to-end training \cite{wang, lrsw}.
In terms of supervision \cite{instance_discr, cpc, contr_loss_chen, momentum_contrast, densepretrain1, wang} use instance discrimination coupled with strong data augmentation as a self-supervision pretext task while \cite{contr_loss, contr_sup2, contr_loss_sup, kamnitsas} use available annotations to drive sample similarities in a fully supervised setting. \cite{densepretrain1, wang} both use self-supervision to improve performance on dense classification. Our method is a fully supervised contrastive learning scheme specifically designed for dense classification tasks. Also in contrast to other approaches we draw both positive and negative pairs from the same sample using a single model pass and do not require the use of large batch sizes or data queues. 

Several techniques have been proposed \textbf{to improve segmentation performance at object boundaries}. Deep CNNs, with multiple max pooling layers and large overlapping receptive fields exhibit a trade-off between classification and localization performance which can manifest as smooth responses at deep CNN layers \cite{deeplab}. To address this issue, several works \cite{arnab, deeplab, crf2, crf3} have used Conditional Random Fields (CRF) as a post-processing layer of CNN features for dense prediction tasks due to their capacity to model the complex relations between predictions at different locations. Inspired by graphics methods for image rendering, \cite{pointrend} treat segmentation as a rendering problem and introduce a module, operating on top of pre-trained models, that performs point-based segmentation predictions at near-boundary locations, adaptively selected by an iterative subdivision algorithm. Following a multi-task learning approach to improve performance at object boundaries, \cite{lefevre} add a signed distance transform regression loss to the objective function in addition to the dense classification loss, arguing that the additional loss term enhances spatial-awareness and implicitly acts as a regularization term for semantic segmentation. Our method follows a different approach which is to pre-train a CNN using an appropriate pretext task that enhances the network's capacity to disentangle the signal at object boundaries. To our knowledge, there exist no other pre-training methods for improving the performance of pixel level semantic segmentation at the boundaries. Since other methods aim at improving processes that take place after parameter initialization we expect them to be complementary to our pre-training method.

\textbf{Self-attention} was introduced in \cite{aiayn} for the 1D task of machine translation and involves learning useful similarities between features at different positions which are later used to derive weights for averaging them. For vision tasks \cite{nonlocal} were the first to use self-attention as an additional layer to {\it Convolutional Neural Network} (CNN) architectures. For an input map $\mathbfcal{X} \in \mathbb{R}^{H \times W \times D_{in}}$, where $H, W$ are the spatial dimensions and $D_{in}$ is the input number of features, queries, keys and values $\mathbf{q}_{ij} = \mathbf{W_q} \mathbf{x}_{ij} \in \mathbb{R}^{D_{q}}$, $\mathbf{k}_{ij} = \mathbf{W_k} \mathbf{x}_{ij} \in \mathbb{R}^{D_{q}}$, $\mathbf{v}_{ij}=\mathbf{W_v} \mathbf{x}_{ij} \in \mathbb{R}^{D_{out}}$ are calculated as linear transformations of the input at every spatial location. Learnt matrices $\mathbf{W_q}, \mathbf{W_k} \in \mathbb{R}^{D_q \times D_{in}}$, $\mathbf{W_v} \in \mathbb{R}^{D_{out} \times D_{in}}$ control the number of features $D_q$ (queries, keys) and $D_{out}$ (values) respectively. The similarity matrix $\mathbf{S}_{ij} = \mathbfcal{K}_\mathcal{N} \bullet \mathbf{q}_{ij}$ contains the pairwise affinities between location $(i,j)$ and all other locations in the layer's receptive field $\mathcal{N}$. These affinities are subsequently softmax-normalized such that their values sum to one and are used to aggregate the projected inputs (values).

\begin{equation}\label{self-attn}
    \mathbf{y}_{ij} = \sum_{p \in \mathcal{N}}softmax_p(\mathbf{q}_{ij}^T\mathbf{k}_p)\mathbf{v}_p
\end{equation}

The self-attention formulation in eq.\ref{self-attn} is permutation invariant with respect to the input, meaning that any change in the spatial ordering of the inputs does not affect the output. To exploit positional information, which is critical to capture the spatial structures in vision data, \cite{attn_augm_cnn} introduced the use of two dimensional learnt relative positional encodings to remove the permutation invariance property while remaining translation equivariant. A learned relative positional encoding term $\mathbfcal{R}$ is incorporated into the affinities $\mathbf{S}_{ij} = (\mathbfcal{K}_\mathcal{N} + \mathbfcal{R}) \bullet \mathbf{q}_{ij} $, $\mathbf{r}_{ij} \in \mathbb{R}^{D_{q}}$ introducing a prior of where to look at in the receptive field $\mathcal{N}$ relative to location $(i,j)$. 

\begin{equation}\label{self-attn-rel}
    \mathbf{y}_{ij} = \sum_{p \in \mathcal{N}}softmax_p(\mathbf{q}_{ij}^T(\mathbf{k}_p + \mathbf{r}_p))\mathbf{v}_p
\end{equation}

We note that in addition to relative positional encodings, absolute positional encodings have also been employed \cite{aiayn} but their performance has been shown to be inferior in both language \cite{relative_positional_language} and vision tasks \cite{stand_sa}. 
We also note that $\mathcal{N}$ is abstractly defined as the receptive field of the attention mechanism in eqs.\ref{self-attn}, \ref{self-attn-rel}. Early works \cite{aiayn, nonlocal} defined $\mathcal{N}$ as the full extent of the input allowing each location to attend to all other locations. This could be very desirable for tasks that require modelling of long range dependencies, however, it makes the calculation of pair-wise affinities $\mathbfcal{S}$ prohibitive for high resolution images as its complexity is quadratic with respect to the total number of input locations $\mathcal{O}(H^2W^2)$.

To address this problem \cite{stand_sa} restrict the receptive field of the self-attention module to a local $(h \times w)$ region and reduce the computational complexity to $\mathcal{O}(HWhw)$ allowing them to use this module as stand-alone primitive in place of convolution layers to construct deep architectures. The formulation of the constrained self-attention module with relative positional encodings takes the following form:

\begin{equation}\label{self-attn-rel-local}
    \mathbf{y}_{ij} = \sum_{p \in {\mathcal{N}_{h \times w}(ij)}}softmax_p(\mathbf{q}_{ij}^T(\mathbf{k}_p + \mathbf{r}_p))\mathbf{v}_p
\end{equation}

Where $\mathcal{N}_{h \times w}(ij)$ denotes the local $h \times w$ region around location $(i,j)$. 
In practice instead of applying the attention function once with $D_q$ and $D_{out}$ number of features, all queries, keys and values are projected into $n$ different spaces with dimensions $\frac{D_q}{n}$ and $\frac{D_{out}}{n}$ and the self-attention module is applied $n$-times in parallel. The outputs are further concatenated in the features dimension and are potentially linearly projected to obtain the final values.

Our method, inspired by the self-attention module and particularly the work of \cite{stand_sa} calculates a local affinity matrix $\mathbfcal{S}$ using the final layer features of a CNN but rather than using it for feature aggregation we apply supervision directly on the calculated affinities. 
Also, because we are interested only in the total pair-wise affinities we only make use of a single attention head.

\begin{figure*}[!h]
\begin{center}
   \includegraphics[width=1.0\textwidth]{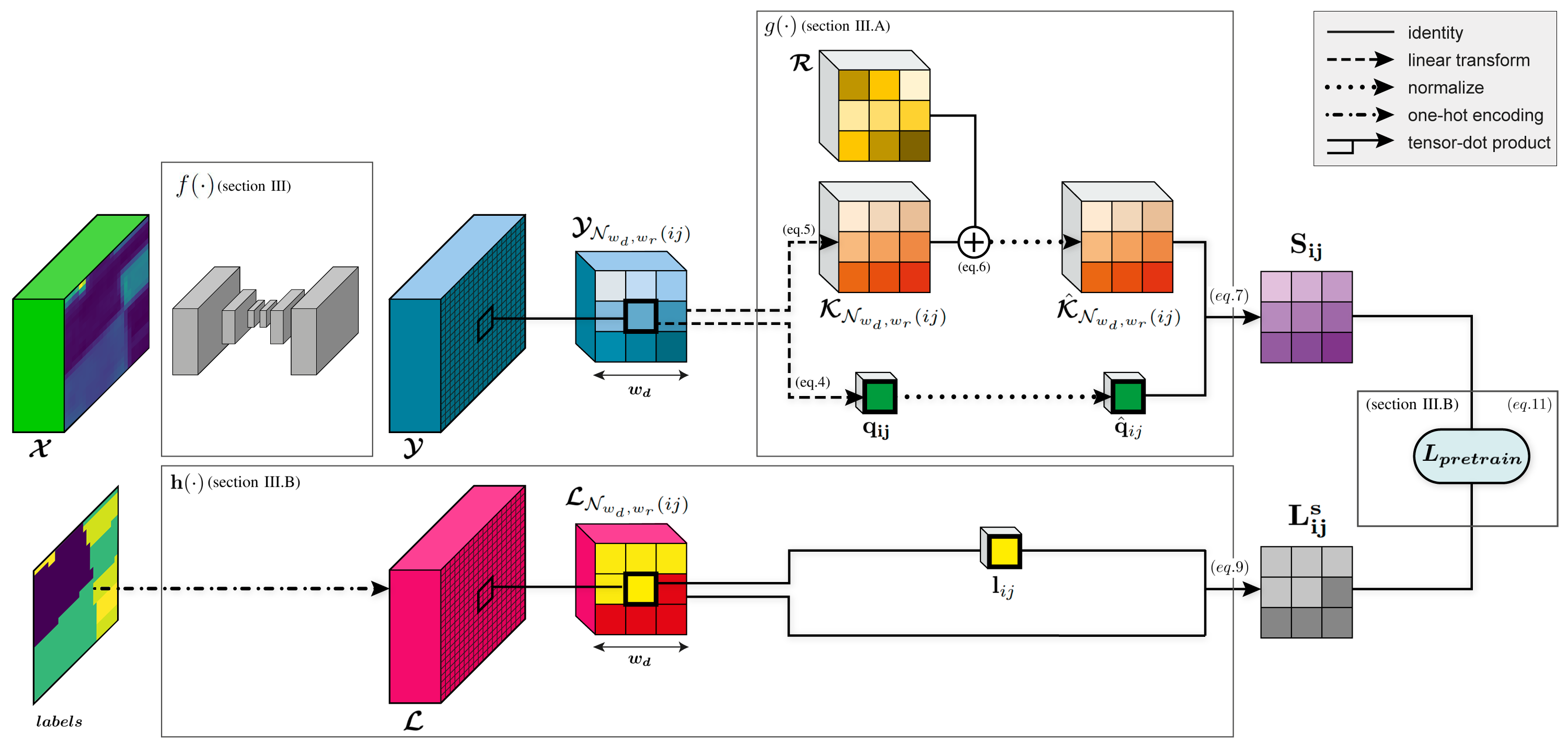}
\end{center}
   \caption{Proposed pre-training scheme using the {\it Context-Self Contrastive Loss} ({\it CSCL}). For a {\it CNN} feature map $\mathbfcal{Y}$ we define a similarity metric $\mathbf{S}_{ij}$ between features extracted at all locations $(i, j)$ and every location in their $w_d$, $w_r$-dilated neighborhood (top branch). Similarly, we derive training labels $\mathbf{L}^s_{ij}$ from dense annotations $\mathbfcal{L}$ for fully supervised training (bottom branch).}
    \label{cscl}
\end{figure*}

\section{Context-Self Contrastive pre-training}\label{proposed_method}

Applied on an N-dimensional embedding space of a CNN, the affinity matrix of a self-attention layer naturally encodes semantic similarity between features extracted at different locations. Using densely annotated data, we derive class agnostic labels for providing full supervision on the values of the similarity matrix and train a network on the task of predicting class agreement between each location paired with all locations in its local neighbourhood. 

In its most general form, the proposed pre-training method includes the following components:
\begin{enumerate}
    \item an encoder function $\mathbfcal{Y} = f(\mathbfcal{X})$ mapping an input tensor $\mathbfcal{X}$ to an output feature map $\mathbfcal{Y}$
    \item a self-similarity function $\mathbfcal{S} = g(\mathbfcal{Y})$
    \item a ground truth generator module $\mathbfcal{L}^\mathbf{s} = h(\mathbfcal{L})$ reformatting dense class annotations $\mathbfcal{L}$ into class-agnostic $\mathbfcal{L}^\mathbf{s}$ 
    \item a pre-train loss function $L_{pretrain}(\mathbfcal{S}, \mathbfcal{L}^s)$
   
\end{enumerate}
In this paper, the encoder function is a deep CNN, $\mathbfcal{Y}$ and $\mathbfcal{L}$ are order-3 tensors while $\mathbfcal{S}$ and $\mathbfcal{L}^\mathbf{s}$ are order-4 tensors. 

\subsection{Similarity function}
Drawing inspiration from \cite{stand_sa} we use a local affinity matrix of embeddings $\mathbfcal{Y}$ as our similarity metric. Formally, for a feature map $\mathbfcal{Y} \in \mathbb{R}^{H \times W \times D_{in}}$ we calculate {\it queries} and {\it keys} tensors as linear projections of the final layer embeddings.
\begin{equation}\label{attnl_q}
    \mathbf{q}_{ij}=\mathbf{W_q} \mathbf{y}_{ij}  \in \mathbb{R}^{D_q}    
\end{equation}
\begin{equation}\label{attnl_k}
    \mathbf{k}_{ij}=\mathbf{W_k} \mathbf{y}_{ij} \in \mathbb{R}^{D_q}    
\end{equation}
Where $\mathbf{W_q}, \mathbf{W_k} \in \mathbb{R}^{D_q \times D_{in}}$ are learnt matrices and $D_{in}, D_q$ define the number of features of the {\it input} and {\it keys}, {\it queries} tensors. In all experiments we use $D_{in}=128$ and $D_q=128$ although similar results are obtained with different choices for these parameters. 

We calculate the similarity tensor $\mathbfcal{S}$ as the tensor-dot product between unit-sphere normalized {\it queries} $\hat{\mathbf{q}_{ij}}=\frac{\mathbf{q}_{ij}}{\|\mathbf{q}_{ij}\|}$ and normalized local {\it keys} which are first augmented by relative positional encodings. The inclusion of relative position information takes place by directly adding $\mathbfcal{R} \in \mathbb{R}^{w_d \times w_d \times D_q}$ to the local keys. $\mathbfcal{R}$ is composed of learnt parameters $\mathbf{R}^H \in \mathbb{R}^{w_d \times \frac{D_q}{2}}$ and $\mathbf{R}^W \in \mathbb{R}^{w_d \times \frac{D_q}{2}}$ representing relative height and width information respectively. We calculate $\mathbf{r}_{ab} = [\mathbf{r}^H_a, \mathbf{r}^W_b]$ for each location $(a,b)$ within the local window. Here $[\cdot, \cdot]$ denotes vector concatenation. 

\begin{equation}\label{add_R}
    \mathbfcal{K^R}_{\mathcal{N}_{w_d, w_r}(ij)}=\mathbfcal{K}_{\mathcal{N}_{w_d, w_r}(ij)}  + \mathbfcal{R}  
\end{equation}

\begin{equation}\label{csl_s}
    \mathbf{S}_{ij}=\hat{\mathbfcal{K^R}}_{\mathcal{N}_{w_d, w_r}(ij)} \bullet \hat{\mathbf{q}_{ij}} \in \mathbb{R}^{w_d \times w_d}    
\end{equation}

$\mathcal{N}_{w_d, w_r}(ij)$ denotes the $(w_d \times w_d)$, $w_r$-dilated neighbourhood centered around location $(i,j)$. For keys normalization we normalize each spatial element separately by its norm. Normalizing both {\it keys} and {\it queries} is a key step for the method to work as not doing so would result in an ever increasing vector norm when training with the objective function presented in eq.\ref{csl_contr_loss}. The process for calculating local affinities is presented schematically at the top branch of Fig.\ref{cscl}. 

\begin{figure}[t]
\centering
\includegraphics[width=0.6\linewidth]{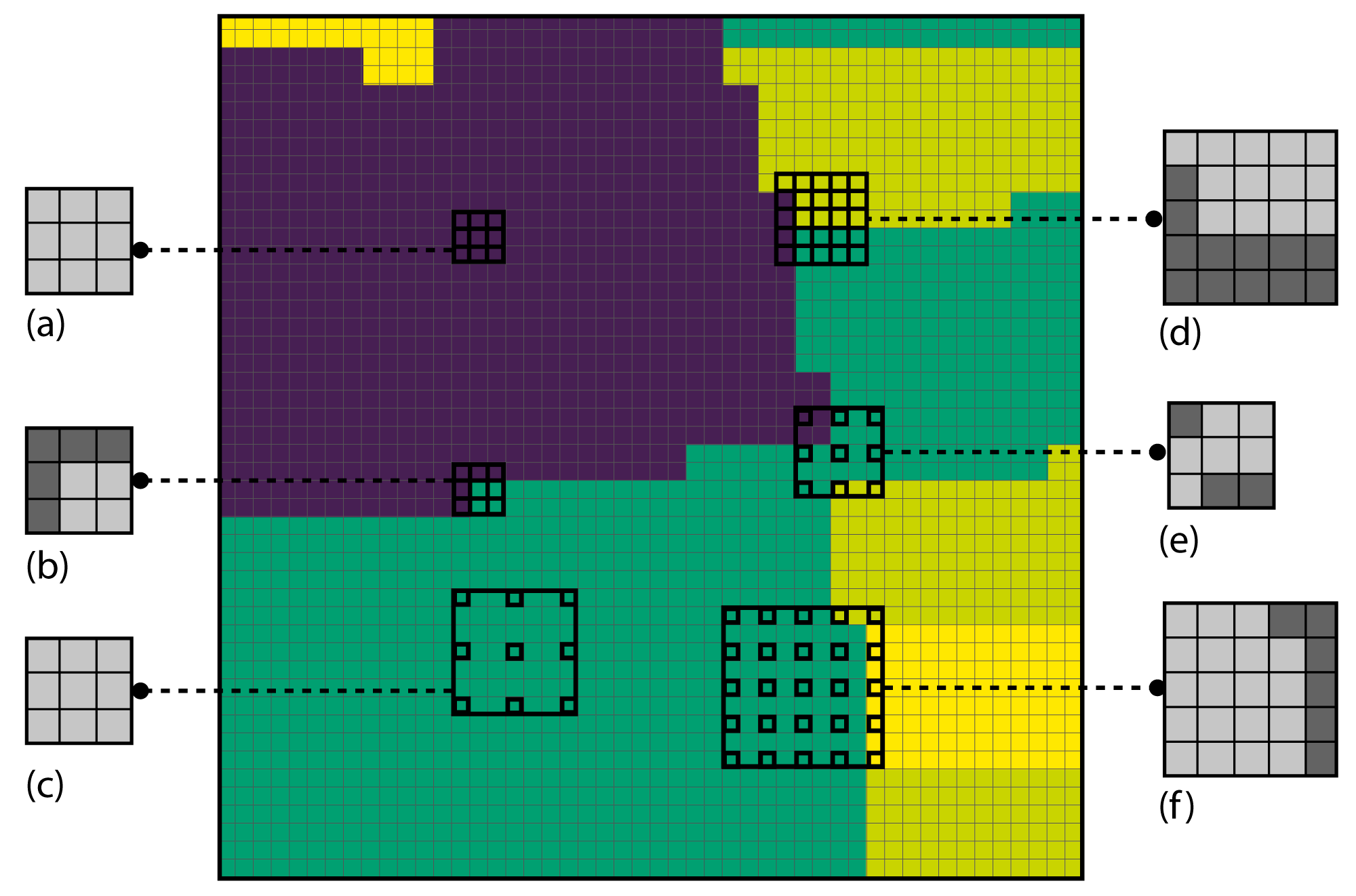} 
\caption{CSCL ground truth generator example on a 2D crop type label map. To generate labels we compare the class at the center location with all locations in the local neighbourhood defined by parameters $w_d, w_r$. We use windows with parameters (a, b) $w_d=3, w_r=1$, (c) $w_d=3, w_r=3$, (d) $w_d=5, w_r=1$, (e) $w_d=3, w_r=2$, (f) $w_d=5, w_r=2$.}
\label{fig:cscl_label_example}
\end{figure}

\subsection{Loss function}

Before defining the pre-training loss function we need to reformat the ground truth data such that they encode the similarity between each location in a 2D lattice and all locations in its local neighbourhood. We start from dense one-hot ground truths $\mathbfcal{L} \in \{0, 1\}^{H \times W \times C}, \sum_k \mathbfcal{L}_{ijk}=1 \: \forall \: i, j$ of spatial extent $H\times W$ and $C$ the number of classes. In essence we need pre-training ground truths $\mathbfcal{L}^s \in \{0, 1\}^{H \times W \times w_d \times w_d}$ to be a binary tensor of the same dimensions as $\mathbfcal{S}$ with values

\begin{equation}
  l^s_{ijkm} =
    \begin{cases}
      1 &, \mathbf{l}_{i, j}=\mathbf{l}_{i-w_r(c-k), j-w_r(c-m)}\\
      0 &, otherwise\\
    \end{cases}       
\end{equation}

where $c=\floor{\frac{w_d}{2}}$ is the index for the center location of the local window and $(k, m) \in \{0,...,w_d-1\}^2$ define a location within the spatial extent of the local window.  
Formally, we obtain suitable labels that encode class agreement between each location $(i,j)$ and its $(w_d \times w_d)$, $w_r$-dilated neighbourhood by application of the tensor-dot product.
\begin{equation}\label{attnl_s}
    \mathbf{L}^{s}_{ij} = \mathbfcal{L}_{\mathcal{N}_{w_d, w_r}(ij)} \bullet \mathbf{l}_{ij} \in \{0,1\}^{w_d \times w_d}    
\end{equation}
$\mathcal{N}_{w_d, w_r}(ij)$ denotes the $(w_d \times w_d)$, $w_r$-dilated neighbourhood centered around location $(i,j)$. 
The process is shown at the bottom branch of Fig.\ref{cscl} while an example of generated labels for various size and dilation parameters is shown in Fig.\ref{fig:cscl_label_example}. 

For $L_{pretrain}$ we use a contrastive loss function extending \cite{contr_loss} to a dense embedding space and using cosine similarities between elements of $\mathbfcal{Q}$ and $\mathbfcal{K}$ instead of Euclidean distances. In order to mask supervision at the zero-padded regions, at the centers of extracted windows and any other location we might want to exclude from learning, e.g. background class, we use a mask tensor $\mathbfcal{M} \in \{0,1\}^{H \times W \times w_d \times w_d}$ of the same dimensions as $\mathbfcal{S}$, $\mathbfcal{L}^\mathbf{s}$.

\begin{equation}\label{csl_contr_loss}
\begin{split}
    L_{pretrain}=-\frac{1}{\sum \mathbfcal{M} }\sum (\mathbfcal{M} \odot (\mathbfcal{L}^\mathbf{s} \odot \mathbfcal{S} + (1 - \mathbfcal{L}^\mathbf{s})\\ 
    \odot \min (0, m-\mathbfcal{S}))) 
\end{split}
\end{equation}

Here, summation is over all dimensions of $\mathbfcal{M}, \mathbfcal{L}^\mathbf{s}, \mathbfcal{S}$ with indices not shown to avoid cluttering. $m$ is a scalar hyperparameter used to exclude easy negatives. With $m>0$ the cosine similarities for easily recognised dissimilar pairs that are smaller than $m$ and have no contribution to the total loss. Several studies have highlighted the importance of hard negative pairs in contrastive learning \cite{contr_loss, hardneg1, hardneg2}. In our method all negative pairs originate from nearby locations in the same training sample featuring high overlaps in their receptive fields. As such there is no possibility for easy negatives and we can discard the $m$ parameter originally employed to exclude the contribution of these pairs. For all our experiments we set $m=0$ and weight the contribution of positive pairs by $\lambda$ leading to the following equivalent expression.
\begin{equation}\label{csl_contr_loss}
    L_{pretrain}=-\frac{1}{\sum \mathbfcal{M} }\sum (\mathbfcal{M} \odot ((\lambda + 1)\mathbfcal{L}^\mathbf{s} - 1) \odot \mathbfcal{S})
\end{equation}

\subsection{Interpreting the sliding window parameters}
\label{sec:slide_params}
\begin{figure}[!tbp]
\centering
\includegraphics[width=0.35\textwidth]{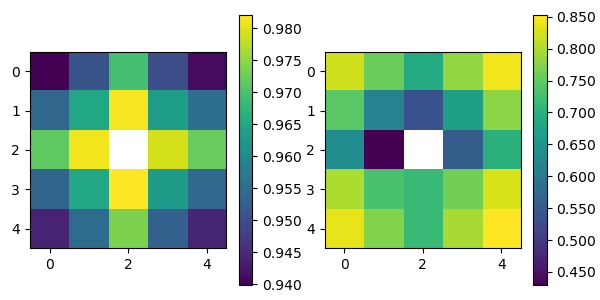}
\caption{Mean prediction accuracy for positive (left) and negative (right) pairs w.r.t location in the sliding window ($w_d=5$). The prediction accuracy drops for positive and increases for negative pairs with increasing distance from the center.}

\label{fig:acc_per_loc}
\end{figure}

We note two ways in which the values of the sliding window parameters $w_d$, $w_r$ affect training. First, they introduce a trade-off between the difficulty of fitting positive and negative pairs. Small $w_d$, $w_r$ values result in comparing locations that are closer together making it easier to recognise positive pairs and harder for negative pairs, the other way being the case for large window size and dilation values.

This can be seen in Fig.\ref{fig:acc_per_loc} where we plot the mean prediction accuracy w.r.t. the location on the sliding window. We assume a positive prediction if the calculated affinity is greater than a specified threshold. We note how the accuracy is shown to be lower for positive pairs the farther away from the center of the sliding window and larger for negative pairs. 

Secondly, they affect the statistics of the generated labels. Small $w_d$, $w_r$ values results in comparing locations that are more likely to belong to the same object resulting in a higher proportion of positive pairs. Class imbalance in class-agnostic ground truths - in favour of positive pairs - has been identified as a key issue. This effect is more pronounced for high resolution outputs with fewer distinct objects, in which case
 
we have found it beneficial to increase $w_d$, $w_r$ and reduce $\lambda$. 
We note the option of applying eqs.(\ref{csl_s}, \ref{attnl_s}) at a stride $w_s$ to reduce the amount of computation in high resolution outputs. This subsampling reduces the spatial dimensions of the feature map, and by extent the attention complexity, by a factor of $w_s^2$ to $\mathcal{O}(\frac{HWhw}{w_s^2})$ compared to the non-strided case. Finally, for some special case of parameters $w_r>1$, $w_s>1$ some locations $(i, j)$ are never used for calculating $\mathbfcal{S}$, facilitating a more efficient implementation. This is further elaborated in Appendix \ref{appendix_cscl_implementation}.

\begin{table*}[!ht]
\begin{center}
\begin{tabular}{|c|c|c|c|c|c|c|c|c|c|c|c|c|}
\hline 
country & $A_{AOI}$ (ha) & $N_{p}$ & $N_{s}$ & $S_{s}$ & $N_{y}$ & $N_{b}$ & $N_{T}$ & $CC_{max}$ & $N_{cl}$ & dense an. & ids an. & SR\\
\hline \hline
France \cite{garnot2019satellite} & 1.21M & 200k & 200k & 32 & 1 & 10 & 24 & - & 20 & - &- &- \\
France \cite{Ruwurm2} $\dagger$ & 2.7M & 608k & 608k  & - &  1 & 13 & 51-102 & $100\%$ & 9 & - & - & -\\
France \cite{Ruwurm2} $\ddagger$ & 2.7M & 608k & 608k  & - & 1 & 10 & 53 & $80\%$ & 9 & - & - & -\\
Austria \cite{timesen2crop} $\star$ & 8.4M & - & 1.2M & - & 1 & 9 & 26-37 & $80\%$ & 16 & - & - & -\\
Ghana \cite{Rustowicz2019SemanticSO} & 2.96M & 4.4k & 4k & 64 & 1 & 10 & 25-100 & $10\%$ & 4 & \checkmark & -& -\\
S.Sudan \cite{Rustowicz2019SemanticSO} & 118M & 837 & 500 & 64 & 1 & 10 & 25-100 & $10\%$ & 4 & \checkmark & -& -\\
Germany \cite{Ru_wurm_2018} & 428k & 137k & 114k & 24 & 2 & 13 & 2-51 & $80\%$ & 17 &  \checkmark & -& -\\
France \cite{garnot_iccv} & 400k & 124k & 2.5k & 128 & 1 & 10 & 33-61 & - & 19 &  \checkmark &\checkmark & -\\
\hline
France (T31TFM-1618)  & 1.21M & 575k & 140k & 48 & 3 & 13 & 14-33 & $70\%$ & 20  & \checkmark & \checkmark  & \checkmark \\
\hline
\end{tabular}
\end{center}
\caption{Large scale datasets of SITS for crop type recognition. $A_{AOI}$: AOI size in {\it hectares}, $N_{p}$: number of parcels in AOI, $N_{s}$: number of samples in the dataset, $S_{s}$: sample size assuming square samples (only for pixel-based datasets), $N_{y}$: number of growing seasons in dataset, $N_{b}$: number of satellite image bands, $N_T$: number of observations for each growing season, $CC_{max}$: {\it maximum} cloud coverage, $N_{cl}$: number of crop types, {\it dense annt.}: dense annotations included, {\it ids annt.}: parcel ids included, {\it SR}: labels at higher resolution than inputs. \cite{Rustowicz2019SemanticSO} contain data from additional satellites, only S2 bands are mentionned here. Input data are: timeseries of mean values over the extent of an agricultural field $\dagger$, $\ddagger$, sampled per pixel timeseries $\star$. S2 images are provided at L1C $\dagger$ and L2A (53 observations on average) $\ddagger$. $N_p, N_s, N_{cl}$ attributes refer only to the subset of data used in the experiments section of each respective publication. For example, out of the 349 RPG codes, our dataset contains 166 different classes that are found in the AOI. From these, only 20 classes were used in experiments (section \ref{experiments}).}
\label{dataset_comparison}
\end{table*}

\begin{figure*}[!t]
\begin{center}
\includegraphics[width=\textwidth, trim={0 1cm 0 0}]{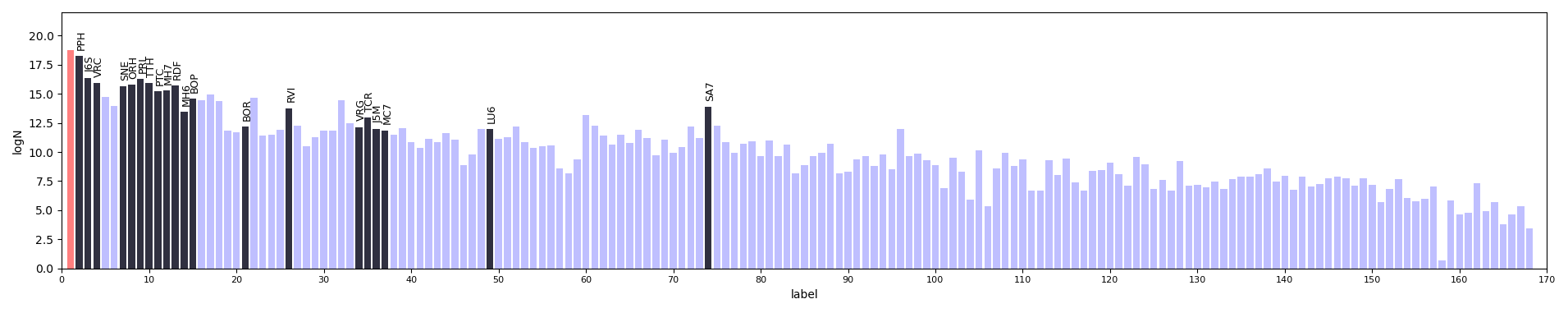}
\end{center}
  \caption{Pixel counts for all 166 crop types available in the T31TFM-1618 dataset. Red column corresponds to background class. Black columns indicate the 20 classes used in experiments.}
\label{fig:all_labels}
\end{figure*}

\begin{figure}[!t]
\begin{center}
\includegraphics[width=0.5\textwidth, trim={70 0 40 0}]{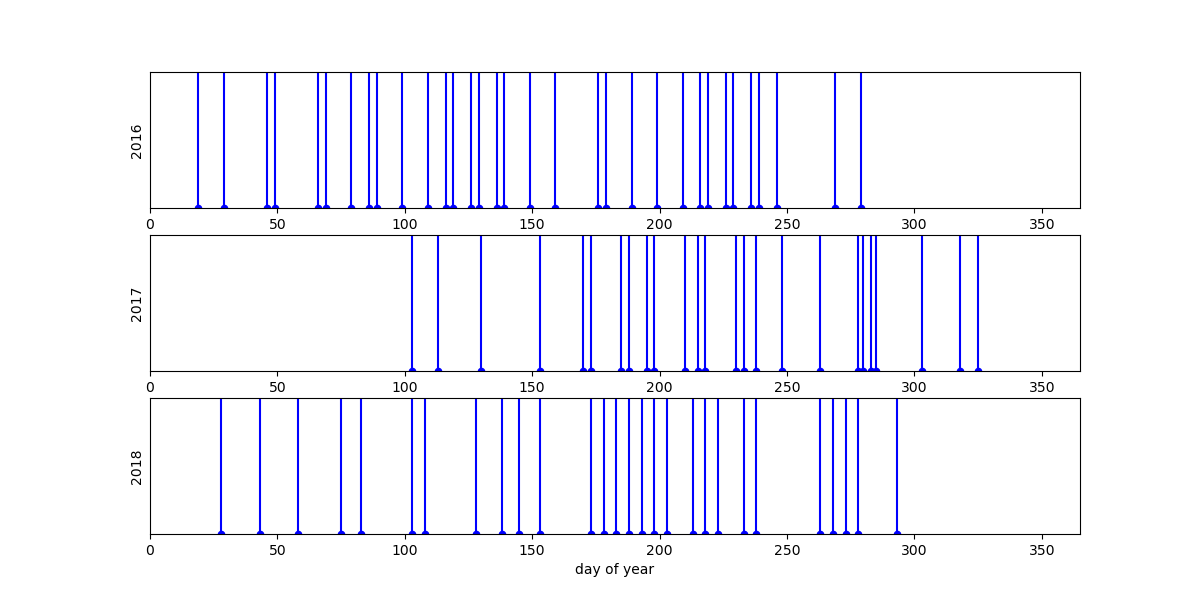}
\end{center}
  \caption{Day of year (doy) of satellite observations included in the T31TFM-1618 dataset.}
\label{fig:doy}
\end{figure}

\begin{figure}[!t]
\begin{center}
\includegraphics[width=0.5\textwidth]{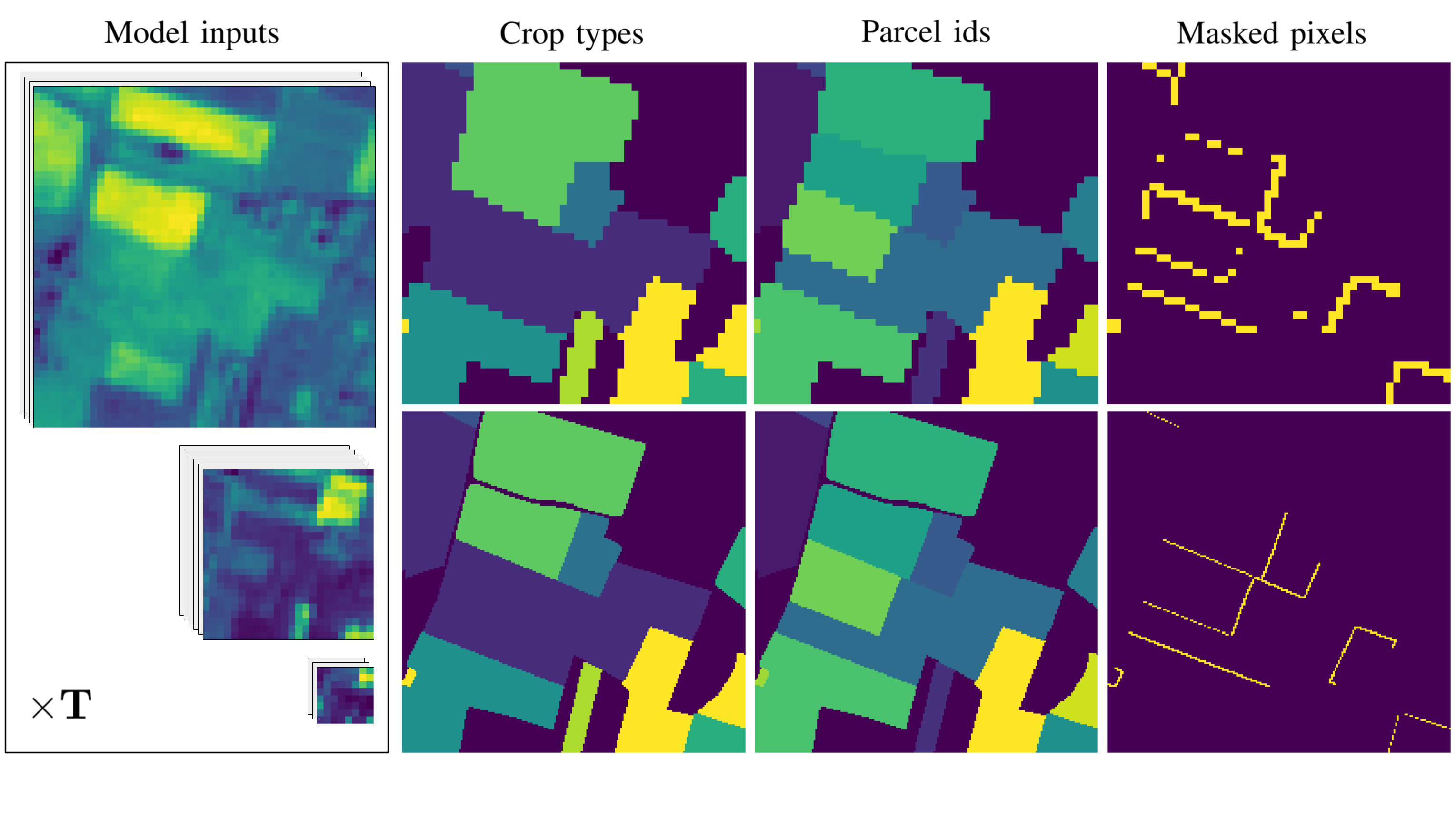}
\end{center}
  \caption{Example of data included in the T31TFM-1618 dataset. (left) Satellite image inputs. We retain four 10m bands, six 20m bands and three 60m bands for variable number of time steps {\it T} depending on location. (right) Ground truth data. From left to right we show crop types, parcel ids and doubly assigned pixels. The same ground truth data are plotted at 10m (top row) and 2.5m (bottom row) resolutions.}
\label{fig:data_sample}
\end{figure}

\section{Experiments}\label{experiments}

\subsection{T31TFM-1618 dataset for crop type segmentation}\label{dataset}
\label{sec:crop_type_data}

For generating our densely annotated crop segmentation dataset we are using ground truth crop type data available in the records of the {\it French Land Parcel Identification System} (RPG) \footnote{https://www.data.gouv.fr/en/datasets/registre-parcellaire-graphique-rpg-contours-des-parcelles-et-ilots-culturaux-et-leur-groupe-de-cultures-majoritaire. In total, the RGP includes approximately $10M$ parcels per year} in the form of parcel geometries. As a source of satellite images we use S2 products available through the {\it Copernicus Open Access Hub} (COAH) \footnote{https://scihub.copernicus.eu/}. The new dataset which we will refer to as T31TFM-1618 covers the full extent of the {\it T31TFM} S2 tile  in France for the years 2016-18. This is the same AOI as \cite{garnot2019satellite}, however, we extend the time period of interest to three years, with separate ground truths provided for each year, and provide dense annotations for crop types and parcel identities. Our AOI also has some overlap with the AOI of \cite{garnot_iccv} which is concurrent work to ours. In Table \ref{dataset_comparison} we compare our dataset with other large scale crop type identification datasets. The T31TFM-1618 dataset is the largest, to our knowledge, publicly available dataset for crop type semantic segmentation. Its AOI covers the full extent of a densely cultivated S2 tile and by including three years it spans the longest time period among similar datasets. In total, 575k agricultural parcels are used for ground truth data. These are uniformly distributed within the 3-year POI, with approximately 192k agricultural parcels used to derive the ground truth data for each year. Additionally, it includes dense annotations for parcel identities enabling the training of object detection and instance segmentation models. In Fig.\ref{fig:all_labels} we show pixel counts per label type for all labels. Days in the form of "day-of-the-year" ({\it doy}) for the observations included in the dataset are shown in Fig.\ref{fig:doy} separate for each year.  

The automated process for generating the dataset includes the following steps: defining an AOI and a POI, downloading all relevant satellite images, extracting and rasterizing all relevant ground truth data, grouping and sorting satellite images to create timeseries objects and matching ground truths with timeseries by location. As a final step, we create {\it boolean} masks for regions that should be masked out during training and evaluation. These include non-agricultural regions, crop types we do not wish to include as well as pixels contained to more than one parcels due to geocoding errors. Samples that contain no classes other than the background class are removed from the training and evaluation sets.

We split inputs into 480m square regions corresponding to $48\times48$ pixels for the highest resolution satellite band. During the ground truth rasterization step we need to make a choice for the size of the grid. We proceed with the rasterization of ground truth polygons using two different grid sizes. A grid size of 10m matches the maximum resolution of S2 images. In this manner, ground truth samples consist of $48\times48$ pixels and each pixel from the highest resolution S2 bands corresponds to one pixel from the raster ground truth data. Additionally, we use a 2.5m grid size to extract the $\times 4$-super-resolution ground truths which leads to a $4 \times 4$ label map per pixel for the highest resolution S2 bands. Super-resolution patches consist of $192\times192$ pixels. Example ground truth data rasterized using 10m and 2.5m grids are presented in Fig.\ref{fig:data_sample}. 
To facilitate further research, we make the T31TFM-1618 dataset publicly available together with evaluation codes. 

\subsection{Network architectures}
\label{sec:crop_type_network}

\begin{figure*}[!ht]
\centering
\includegraphics[width=\textwidth, trim=0 30 0 30, clip]{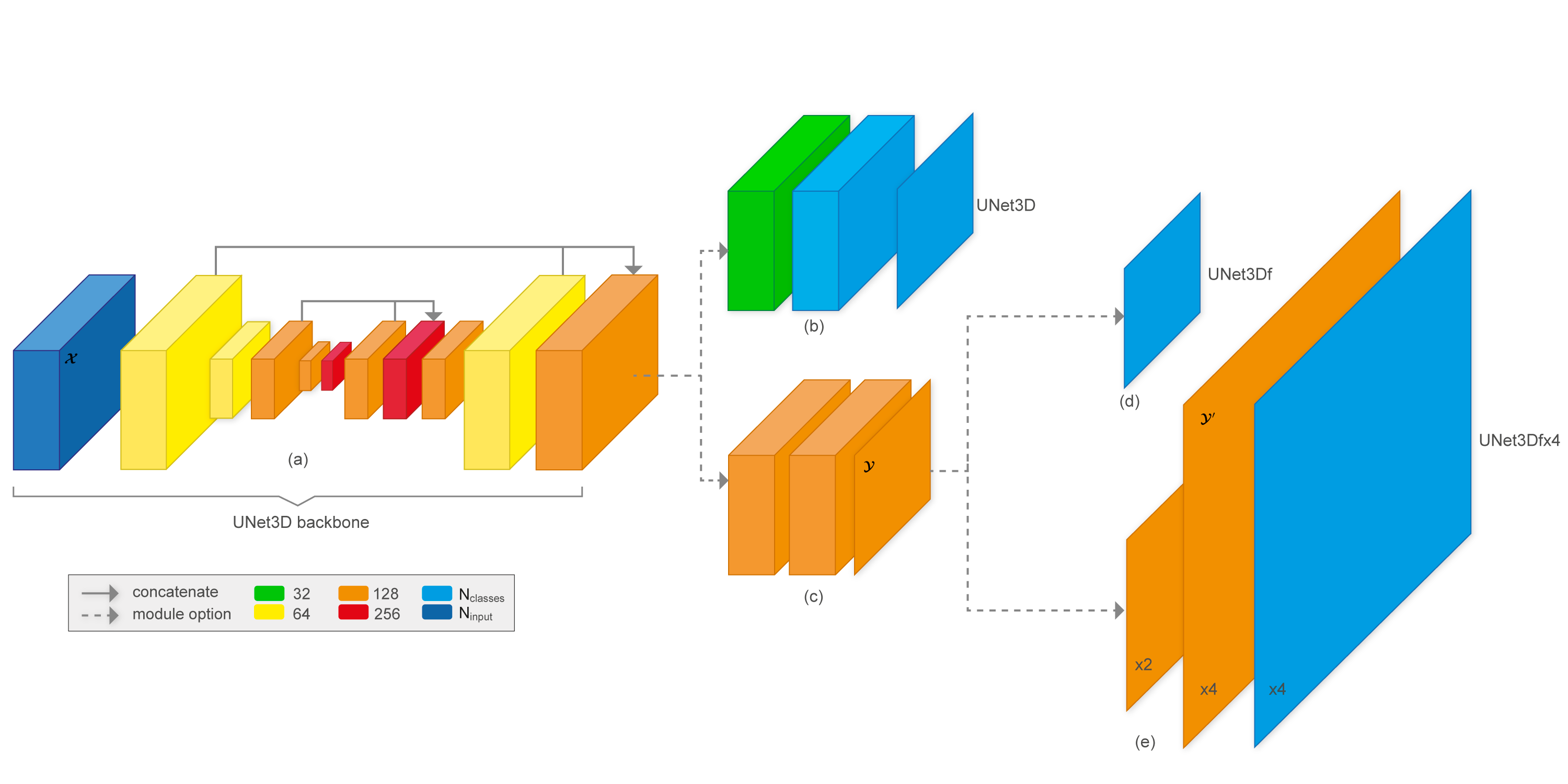}
\caption{UNET3D ($a \rightarrow b$), UNET3Df ($a \rightarrow c \rightarrow d$) and UNET3Df-x4 ($a \rightarrow c \rightarrow e$) models. All models share the same backbone. Dashed lines indicate how different combinations of components lead to different architectures. Box size represents spatial-temporal dimensions, feature dimensions are shown by color coding. Indicators $\times2$, $\times4$ show super-resolution feature maps which are not drawn in scale.}
\label{fig:unet3d}
\end{figure*}

All models explored here learn a mapping from image timeseries to a dense label map , i.e. $f: \mathbb{R}^{T \times H \times W \times D} \rightarrow \mathbb{R}^{H \times W \times C}$ where $T$ is the length of the timeseries, $D$ equals the number of bands in a multispectral satellite image plus a spatially-constant {\it doy} feature and $C$ is the number of crop types. 

In the experiments that follow we use the backbone architectures from \cite{Ru_wurm_2018, Rustowicz2019SemanticSO}. We have modified the {\it UNET3D} architecture \cite{3dunet} to make it more suitable for our contrastive pre-training scheme. Towards this goal we only modified network components after the last residual connection which leads to a feature map of size $(T \times H \times W \times 128)$. In the subsequent layers of UNET3D the number of channels is further reduced to $32$ and $C$ while all spatial-temporal dimensions are not affected. Finally, the classifier layer collapses the temporal dimension resulting to an output logits map of size $(H \times W \times C)$. The issue we encounter here is that the final 2D embedding space contains as few as $C$ channels which can be small to encode meaningful similarities between locations in eq.\ref{csl_contr_loss}. We thus aim to construct a richer final 2D embedding space. In UNET3Df we maintain the number of features from the last unaffected feature map to $128$ for the following two layers and then collapse the temporal dimension. This way the final feature map $y \in \mathbb{R}^{H \times W \times 128}$ contains $128$ features ($128 > C$) which is typical of the number of features used in deep contrastive learning \cite{isola, momentum_contrast, momentum_contrast2}. To obtain dense logits we transform $\mathbfcal{Y}$ into a feature map with $C$ number of features by use of a linear classifier layer.

A schematic comparison between the UNET3D and the modified UNET3Df architectures is presented in Fig.\ref{fig:unet3d}. 

Our super-resolution network builds on the UNET3Df architecture without the classifier layer. The final 2D embedding space $\mathbfcal{Y}$ is further processed by two 2D upsampling blocks each consisting of a deconvolution layer \cite{fcn} with kernel size $3\times 3$ and stride $2$, a {\it Batch Normalization} (BN) layer \cite{batch_norm} and {\it Leaky ReLu} activation \cite{leaky_relu}. We denote $\mathbfcal{Y}' \in \mathbb{R}^{4H \times 4W \times 128}$ the feature map after the two upsampling layers. For CSCL pre-training we now use $\mathbfcal{Y}'$ in the same manner that we used $\mathbfcal{Y}$ for the base resolution case. For semantic segmentation training we apply a linear classifier in every spatial location leading to a $C$-dimensional map of logits. 
A schematic representation of the super-resolution component and its relationship with the UNET3Df architecture is presented in Fig.\ref{fig:unet3d}.

\subsection{Implementation details}
\label{sec:implementation_details}

{\bf Datasets. }
We split the T31TFM-1618 dataset into training and evaluation sets consisting of approximately 40k and 6.5k samples per year using {\it DeepSatData} \cite{deepsatdata}. In total, for all three years, our dataset contains $120,438$ training and $18,749$ evaluation samples with no overlapping locations. We do not apply any label grouping but rather select classes directly from the $166$ RPG codes found in the AOI such that there are at least $20,000$ such parcels in the train set and $2,000$ in the evaluation set for each year resulting in $C=20$ classes in total. Pixel counts for these classes are presented in the black columns of Fig.\ref{fig:all_labels}. From publicly available crop segmentation datasets we make use of the dataset presented in \cite{Ru_wurm_2018} which covers an AOI of $102km \times 42km$ north of Munich, Germany containing $C=17$ classes. We train all models using the provided train split and report results for the evaluation split for a fair comparison with literature.

{\bf Evaluation protocol. }
Our main goal is to test the benefit of the proposed pre-training method as an alternative to random initialization for crop type semantic segmentation. All experiments dedicated to this purpose follow the same two step protocol:
\begin{enumerate}
    \item we obtain the baseline performance for a model of choice and dataset by training the network to convergence starting from random initialization of model parameters
    \item we pre-train the same model using the methodology presented in section \ref{proposed_method} starting from random initialization of model parameters. The pre-trained model is subsequently used as the initialization point for supervised end-to-end training following the same training schedule and hyperparameters as step 1.
\end{enumerate}
Presented evaluation metrics are pixel-level {\it overall accuracy} (averaged over pixels) and {\it mIoU}, {\it macro F1} scores (averaged over classes) in the {\it evaluation} sets for Germany and France. We run all experiments five times at base resolution and three times at super-resolution and report the mean value and the $95\%$ confidence interval around the mean for each metric.

\begin{table*}[!ht]
\begin{center}
\begin{tabular}{|c||c|c|c||c|c|c|}%c|c|}
\cline{2-7}
\multicolumn{1}{c|}{} & \multicolumn{3}{|c||}{Germany} & \multicolumn{3}{|c|}{France} \\
\hline
model & Acc. & mIoU & F1 & Acc. & mIoU & F1\\
\hline \hline
BiCGRU \cite{Ru_wurm_2018} & 0.897 & - &  0.831 & $88.4 \pm 0.1$ & $57.3 \pm 0.1$ & $68.7 \pm 0.1$ \\
UNET3D \cite{Rustowicz2019SemanticSO} & $91.4 \pm 0.2$ & $72.5 \pm 0.1$ & $82.3 \pm 0.2$ & $88.2 \pm 0.1$ & $57.1 \pm 0.2$ & $68.6 \pm 0.2$ \\ 
\hline
UNET2D-CLSTM \cite{Rustowicz2019SemanticSO} & $91.7 \pm 0.2$ & $74.2 \pm 0.2$ & $84.6 \pm 0.2$ & $89.0 \pm 0.2$ & $58.3 \pm 0.2$ & $68.9 \pm 0.2$\\ 
UNET2D-CLSTM-CSCL & $\mathbf{93.2 \pm 0.2}$ & $\mathbf{77.6 \pm 0.2}$ & $\mathbf{86.5 \pm 0.2}$ & $\mathbf{89.9 \pm 0.1}$ & $\mathbf{60.0 \pm 0.1}$ & $\mathbf{70.9 \pm 0.2}$ \\ 
\hline
UNET3Df & $91.6 \pm 0.2$ & $73.2 \pm 0.1$  & $83.0 \pm 0.1$ & $88.5 \pm 0.1$ & $57.2 \pm 0.2$ & $68.8 \pm 0.3$\\
UNET3Df-CSCL &  $\mathbf{92.6 \pm 0.2}$ & $\mathbf{75.9 \pm 0.2}$ &  $\mathbf{84.8 \pm 0.2}$ & $\mathbf{89.3 \pm 0.1}$ & $\mathbf{59.8 \pm 0.1}$ & $\mathbf{70.4 \pm 0.2}$ \\
\hline
% {\bf Ours ViT-like (no pretrain)} & $\mathbf{95.7} & $\mathbf{85.8} & $\mathbf{91.1} & - & $\mathbf{61.8} & - \\
% \hline

\end{tabular}
\end{center}
\caption{Model comparison for Germany and France. Presented metrics are mean and $95\%$ confidence intervals among five runs. Models from \cite{Rustowicz2019SemanticSO} were reimplemented using provided code as reported results in the original study used a custom evaluation set. A significant improvement over the baselines is achieved through CSCL pre-training using no additional data. }
\label{satellite_results_main}
\end{table*}

\begin{figure*}[!ht]
\centering
\includegraphics[width=\textwidth, trim=450 0 450 0, clip]{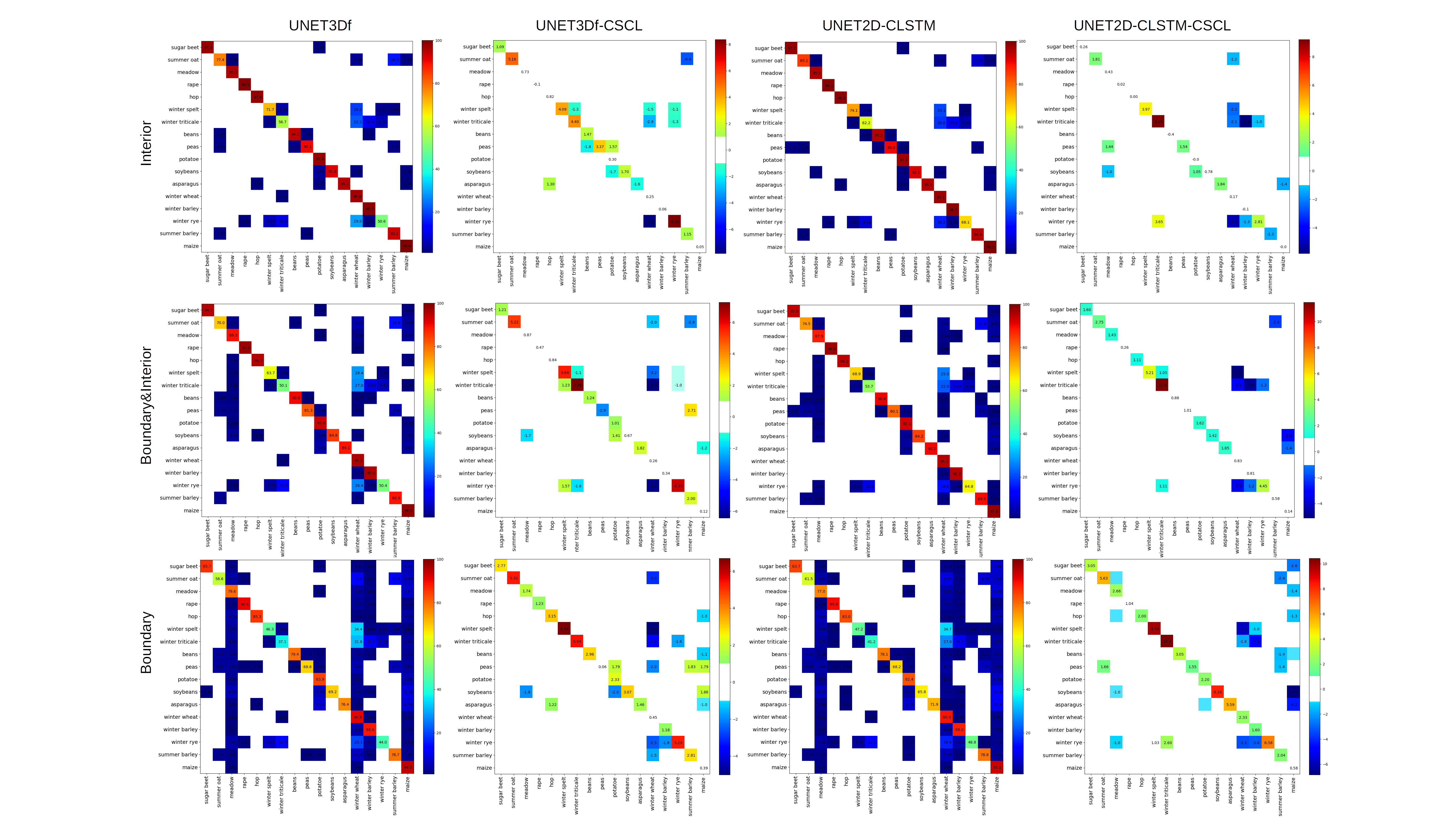}
\caption{Confusion matrices for models trained in Germany. For all baselines we color and explicitly show numerical values for diagonal and all other locations with value $\geq 1.0\%$, all remaining locations are shown in white. For CSCL pre-trained models we display the difference with their respective baseline. We color and explicitly show numerical values for diagonal and all other pairs with absolute difference $\geq 1.0\%$, all other locations are shown in white.}
\label{fig:confmat_de}
\end{figure*}

\begin{figure*}[!ht]
\centering
\includegraphics[width=\textwidth, trim=50 0 150 0, clip]{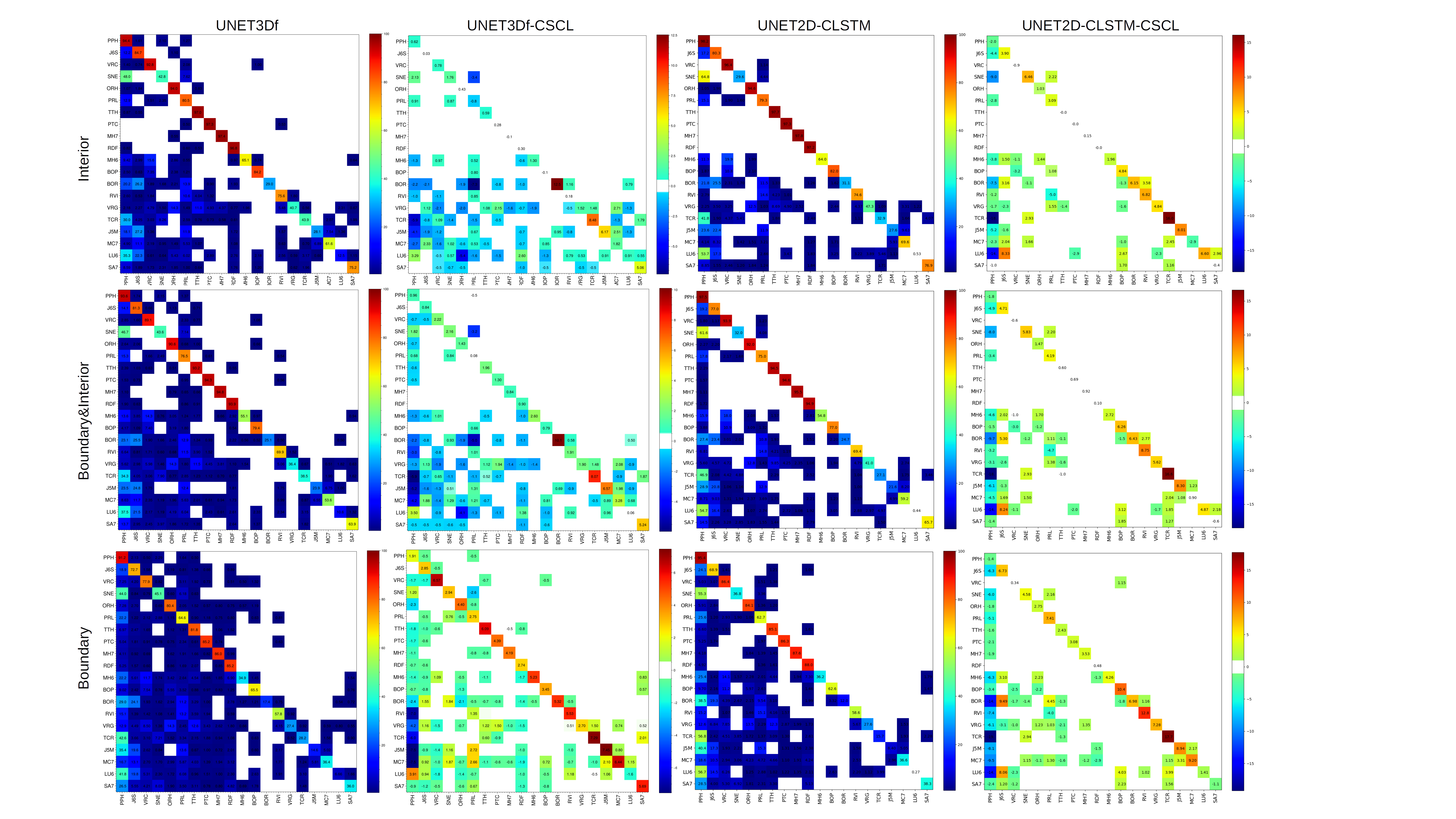}
\caption{Confusion matrices for models trained in France. For all baselines we color and explicitly show numerical values for diagonal and all other locations with value $\geq 1.0\%$, all remaining locations are shown in white. For CSCL pre-trained models we display the difference with their respective baseline. We color and explicitly show numerical values for diagonal and all other pairs with absolute difference $\geq 1.0\%$, all other locations are shown in white.}
\label{fig:confmat_fra}
\end{figure*}

\begin{figure*}[!h]
\centering
\includegraphics[width=0.95\textwidth, trim={0 100 0 300}]{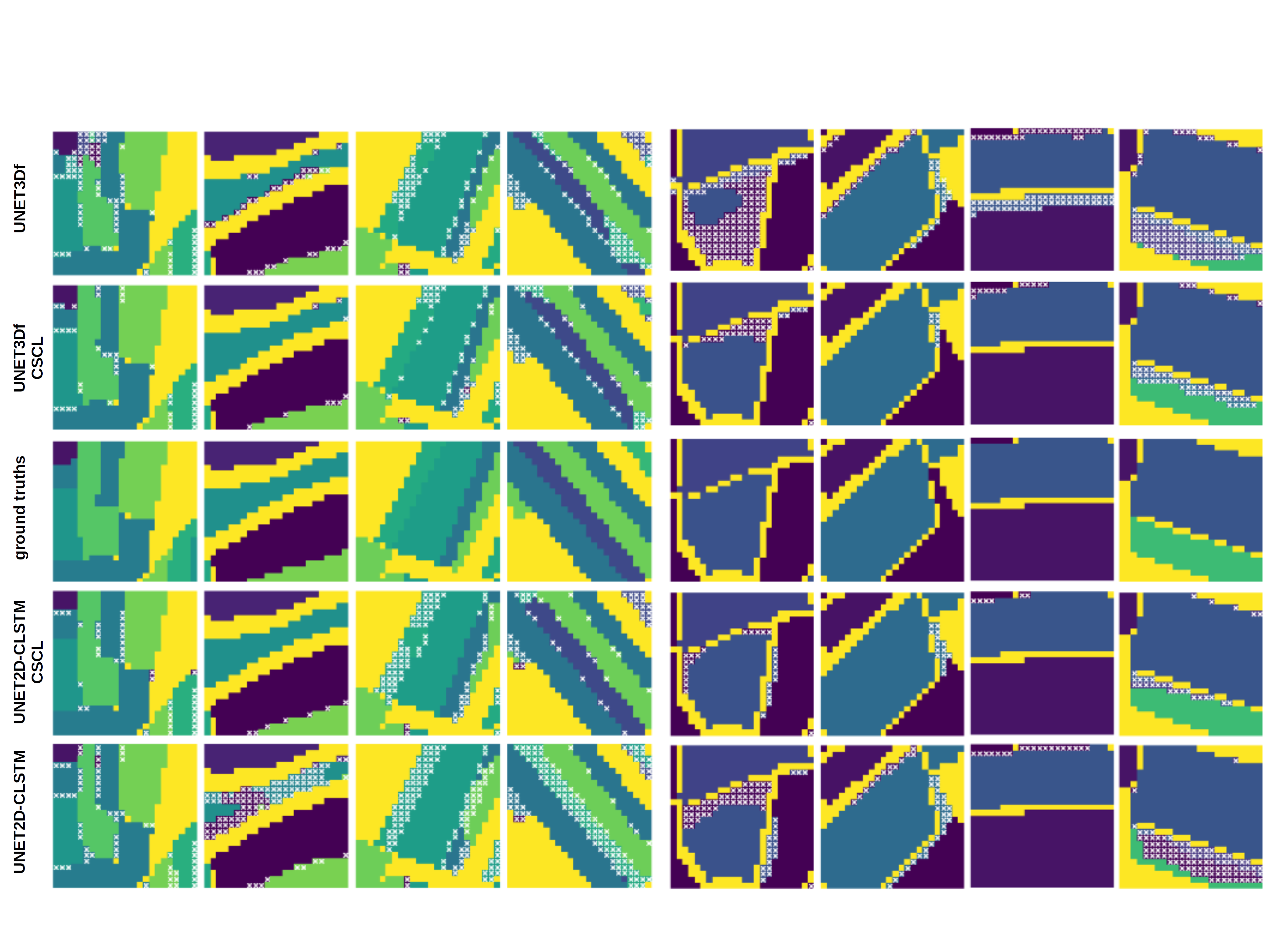}
\caption{Qualitative comparison of models in Germany (left) and France (right). Top to bottom rows: UNET3Df, UNET3Df-CSCL, ground truth crop types,  UNET2D-CLSTM-CSCL, UNET2D-CLSTM. White "x" indicates a false prediction.}
\label{fig:qual_examples_de_fr}
\end{figure*}

{\bf Pre-training and training details. }\label{training_details}
For all experiments, including pre-training and fine-tuning, at base and super-resolution, we use the {\it Adam} optimizer \cite{adam} with initial learning rate $0.0001$ and decay rate $0.975$ applied every second epoch, beta parameters $(0.9, 0.999)$ and batch size $32$. All models are trained for $50$ epochs for France and $150$ epochs for Germany which is enough for the optimization process to converge. We assumed the number of epochs to convergence to be the maximum number of epochs for which evaluation performance did not improve for ten consecutive epochs among all models per country. We attribute the difference in training time to convergence between the two datasets to their relative sizes. Looking at Table \ref{dataset_comparison} our dataset is larger than the Germany dataset both in terms of total number of samples and individual sample size.

Regarding data augmentation, we take advantage of the bilateral symmetry in EO imagery and perform random horizontal and vertical flipping with probability 0.5 in each dimension. We do not perform random cropping but rather use the original patch dimensions per dataset ($24 \times 24$ for Germany and $48 \times 48$ for France). 
For pre-training we use a sliding window with $w_d$=3, $w_r$=1, $w_s$=1 and $\lambda$=0.125. Following the completion of pre-training we discard all components apart from the encoder which is used as an initialization point for fine-tuning. We add a randomly initialized linear classifier mapping encoded features to logits $\mathbfcal{O} \in \mathbb{R}^{H \times W \times C}$, where $C$ is the number of classes, and proceed with end-to-end training of the network using the Masked CE loss function.

\section{Results} 

\subsection{Comparison with state-of-the-art}\label{compare_sota}
We compare our method's performance over random initialization in training recently proposed models using the T31TFM-1618 and German datasets. Table \ref{satellite_results_main} and Figs.\ref{fig:confmat_de}, \ref{fig:confmat_fra} respectively present performance metrics and confusion matrices while Fig.\ref{fig:qual_examples_de_fr} visually compares predictions. Both quantitative and qualitative comparisons indicate a clear advantage of CSCL over the randomly initialized baselines, achieving new state-of-the-art results. 
Quantitatively, we find the absolute {\it mIoU} improvement for UNET2D-CLSTM and UNET3Df to be $+3.4\%$ and $+2.7\%$ in Germany and $+1.7\%$, $+2.6\%$ respectively in France. These differences are much larger than their respective standard deviations shown in Table II for all reported metrics. This suggests that performance improvements can be attributed to our pre-training scheme and not to variations in best model performance among different training sessions. By comparing the relative gains of different metrics, we find the smallest improvements for the {\it overall accuracy}. A possible explanation for this observation could be that {\it overall accuracy}, which is the only metric averaged over pixels, is influenced by the class imbalance in both datasets to a greater extent than the other class-averaged metrics. For example, observing Figs.\ref{fig:all_labels} and \ref{fig:confmat_fra} for France, we find that some of the classes with the highest relative gains (BOR, TCR, J5M) contain relatively few samples. Thus, we expect their effect to be more pronounced in class-averaged rather than pixel-averaged metrics.

\begin{figure}[!h]
    \centering
    \includegraphics[width=\linewidth]{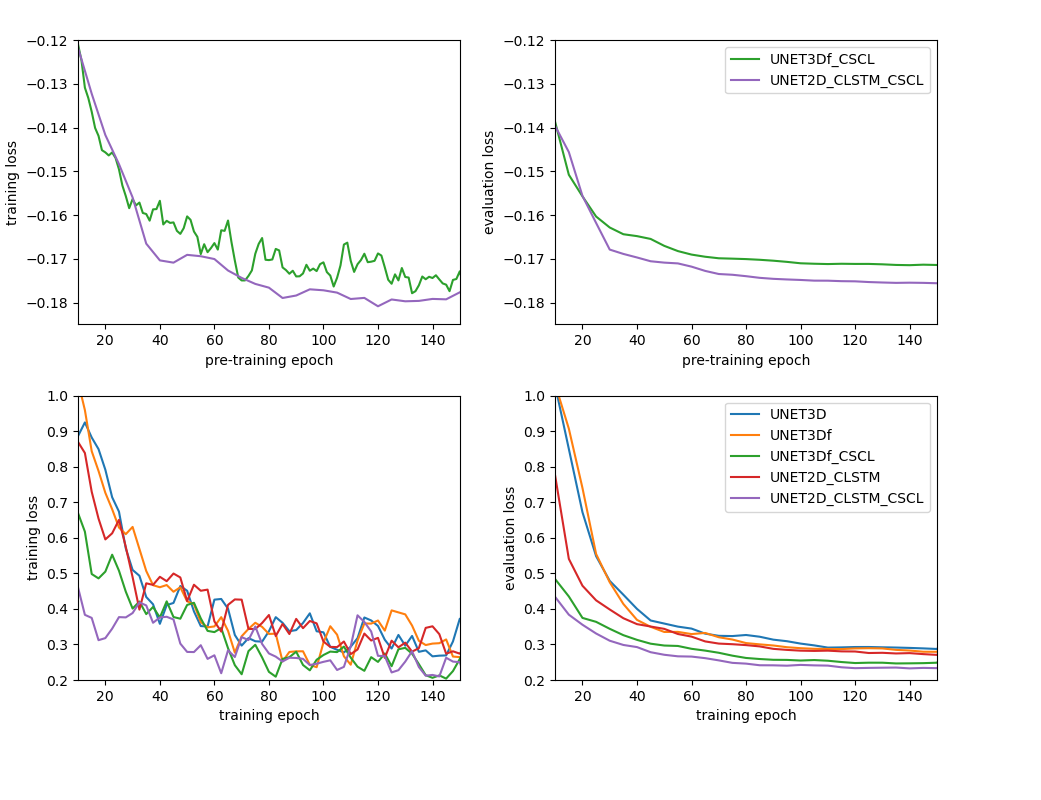}
    \caption{Germany loss curves for CSCL pre-train task (top row) and semantic segmentation training (bottom row). Figures show respective losses calculated using  batches from the training (left column) and the evaluation sets (right column).}
    \label{fig:loss_curves}
\end{figure}

\begin{figure}[!t]
\centering
\includegraphics[width=0.5\textwidth]{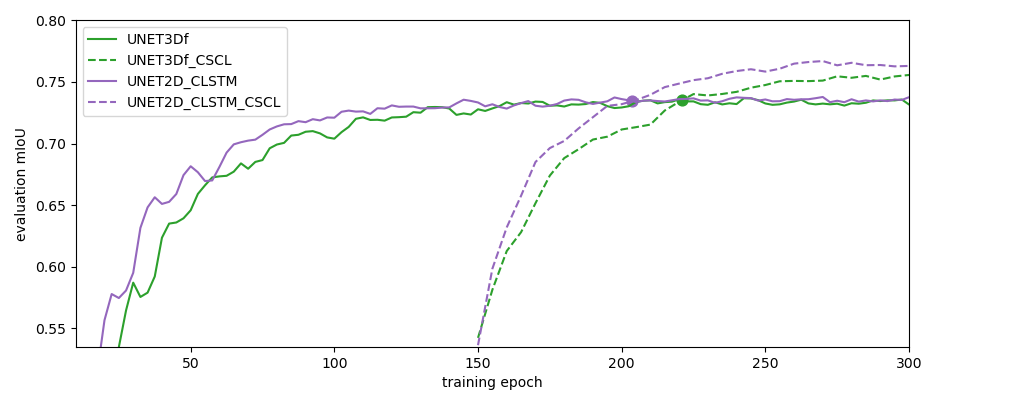}
\caption{Model performance (CSCL vs baselines) in terms of absolute number of training epochs. For pre-trained models we add 150 epochs to the corresponding training time to account for the time spent in pre-training. The points at which pre-trained models surpass the baseline performance are indicated by dots of the same color as the curves representing each architecture.}
\label{fig:pretrain_catchup}
\end{figure}

In Fig.\ref{fig:loss_curves} we present training and evaluation loss curves for Germany. Both training and evaluation losses are lower for the pre-trained models compared to their random initialization counterparts. We observe a higher variation in training rather than evaluation losses, which is reasonable given the smaller sample size of training batches compared to the evaluation set. All losses appear to gradually decrease with more training epochs suggesting stable training dynamics. From Fig.\ref{fig:loss_curves} we point out that the loss for pre-trained models is significantly lower than that of the baselines during the first epochs with baselines slowly reducing the difference towards the end. This profile is expected considering the additional time pre-trained models have spent processing the data. To establish that our baselines are not limited by a short training schedule we train our baselines for 300 epochs in Germany, otherwise following the training schedule presented in section \ref{training_details}. We present evaluation {\it mIoU} curves for these models in Fig.\ref{fig:pretrain_catchup}. In the same figure we show {\it mIoU} curves for the pre-trained models without the extra training time. We assume 150 epochs have already been spent during pre-training, thus we add that amount to the number of epochs for the pre-trained models. Fig.\ref{fig:pretrain_catchup} suggests that despite following an extended training schedule baselines do not reach the performance of the pre-trained models. Furthermore, we can see exactly the point at which the pre-trained models surpass the baselines which is after 53 epochs for UNET2D-CLSTM and 70 epochs for UNET3Df. We also note the existence of a better training schedule for Germany which we mention here for completion. Training for 300 epochs with a learning rate restart at epoch 150 (equivalent to running the schedule presented in section \ref{sec:implementation_details} twice) lead to similar improvements in both baselines and pre-trained models. Following this schedule the final {\it mIoU} performance is (baseline/pre-train) $78.8\%/81.0\%$ for UNET2D-CLSTM and $76.8\%/79.4\%$ for UNET3Df. For France, this strategy did not lead to improved performance.

Finally, it is worth considering why the final performance is significantly lower for our experiments using the T31TFM-1618 compared to the German data. This is mainly attributed to a few bad performing classes as shown in Fig.\ref{fig:confmat_fra}. As a sanity check we run further tests with the label groupings from \cite{garnot2019satellite} who reported {\it overall accuracy} $0.942$ and {\it mIoU} $0.509$ reaching similar performance ($0.963$ and $0.543$) and although not directly comparable it indicates the added difficulty of training in this AOI.

\begin{figure}[!t]
\centering
\includegraphics[width=0.5\textwidth]{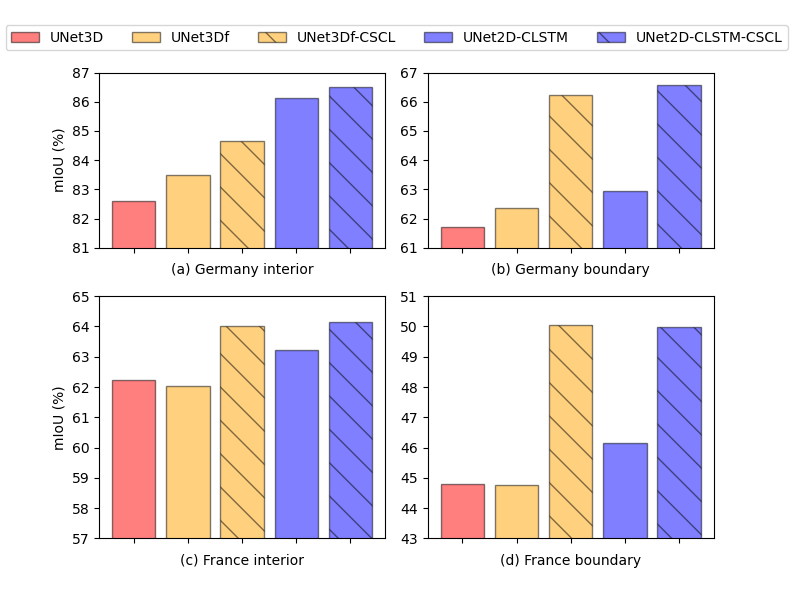}
\caption{Model performance at parcel interior vs boundary locations. We demonstrate that the boost in segmentation performance mostly occurs near the parcel boundaries.}
\label{fig:model_comparison}
\end{figure}

\subsection{Performance at parcel boundaries} \label{boundary_perf}
For our analysis we define a semantic boundary as any location for which not all ground truths in a $3\times3$ local neighbourhood share the same value. By plotting model performance in terms of interior and boundary locations in Fig.\ref{fig:model_comparison} we notice a clear performance drop at boundaries also evidenced by qualitative analysis of results in Fig.\ref{fig:qual_examples_de_fr}. The same conclusion can be reached by inspecting the confusion matrices for interior and boundary points in Figs.\ref{fig:confmat_de}, \ref{fig:confmat_fra}. We also notice smaller but consistent improvements at interior locations for all models, however, using CSCL there is a significant improvement at boundaries over the baselines showcasing the desired effect of the method.

\begin{table*}[!ht]
\begin{center}
\begin{tabular}{|c|c|c|c|c|c|c|c|c|c|}
\hline
model & pre-train & $\mathbfcal{R}=0$ & $\mathbf{W_k}=\mathbf{W_q}$ & $\mathbf{W_k}=\mathbf{W_q}=\mathbf{I}$ & Acc. & mIoU & F1\\
\hline \hline
UNET3Df &  & & & & $88.5 \pm 0.1$ & $57.2 \pm 0.2$ & $68.8 \pm 0.3$\\
UNET3Df-scaled & & & & &  88.40 & 57.70 & 69.20 \\ 
UNET3Df-equal & & & & & 88.60 & 57.93 & 69.60 \\ 
\hline
UNET3Df-CSCL & \checkmark & \checkmark & & & 89.25 & 59.48  & 70.34 \\ 
UNET3Df-CSCL & \checkmark & \checkmark & \checkmark & & 89.30 & 59.22 & 70.23 \\ 
UNET3Df-CSCL & \checkmark & \checkmark & & \checkmark & 89.17 & 59.06 & 69.95 \\ 
\hline
UNET3Df-CSCL & \checkmark & & & & $\mathbf{89.3 \pm 0.1}$ & $\mathbf{59.8 \pm 0.1}$ & $\mathbf{70.4 \pm 0.2}$ \\
UNET3Df-CSCL & \checkmark & & \checkmark & & 89.26 & 59.40  & 70.27 \\ 
UNET3Df-CSCL & \checkmark & & & \checkmark & 89.27 & 59.13 & 70.04 \\
\hline
\end{tabular}
\end{center}
\caption{Results of ablation study in France. (top) Alternative methodologies to pre-training for reducing statistical bias towards interior locations. "UNET3Df-scaled" is initialized randomly and trained such that the loss contribution from interior pixels is scaled down by the ratio (\# boundary pixels)/(\# interior pixels). "UNET3Df-equal" is initialized randomly and trained on an equal number of interior and boundary pixels by randomly masking out a variable number of interior pixel locations for every training sample. (center) Ablation on CSCL components without positional encodings, and (bottom) ablation on CSCL components with positional encodings.}
\label{tab:satellite_ablation}
\end{table*}

\subsection{Ablation study}\label{sat_loss_ablation}
We present an ablation study on CSCL using the T31TFM-1618 dataset. Here we only use the UNET3Df model mainly because it is faster to train compared to UNET2D-CLSTM. 

First, given the large number of interior compared to boundary pixels, we present a {\bf comparison with simpler methods aiming to reduce the statistical bias of the model towards interior pixels}. Towards this goal, we discard pre-training and train directly by scaling - separately for each sample - the loss component from interior pixels by the ratio of $\gamma = \frac{\textrm{num. boundary pixels}}{\textrm{num. interior pixels}}$ (UNET3Df-scaled). In a similar spirit we also train on a balanced number of interior and boundary pixels (UNET3Df-equal). For every training sample we simply mask some interior (or boundary) locations at random such that an equal number of respective locations contribute in the loss function. Results, presented in Table \ref{tab:satellite_ablation}, suggest that scaling down the loss component of interior pixels does improve performance compared to the baseline case ($57.9\%$ vs $57.3\%$ mIoU) but this effect is much less significant compared to the improvements obtained by CSCL ($+0.6\%$ vs $+2.5\%$ mIoU).

Next, we perform an {\bf ablation on the various CSCL components} presented in section \ref{proposed_method}. In particular, we examine the downstream segmentation performance with or without the inclusion of positional encodings in eq. \ref{csl_s}, assuming that learnt parameters $\mathbf{W_q}=\mathbf{W_k}$, i.e. same {\it keys} and {\it queries}, and finally assuming $\mathbf{W_q}=\mathbf{W_k}=\mathbf{I}$, i.e. no feature projection prior to calculating the pairwise affinities. Results are presented in Table \ref{tab:satellite_ablation}. Overall, we observe significant improvements for all variants of the pre-training method compared to random initialization but lower performance compared to the general form of CSCL which includes positional encodings and different projection matrices. 

\begin{table}[!t]
\begin{center}
\begin{tabular}{|l|c|c|c|c|c|c|}
\hline
$\lambda$ & 0.01 & 0.05 & 0.125 & 0.25 & 0.50 & 1.0 \\
\hline \hline
mIoU & 0.551 & 0.556 & \textbf{0.567} & 0.545 & 0.541 & 0.534 \\
\hline
\end{tabular}
\end{center}
\caption{Model {\it mIoU} from varying the weight $\lambda$ of loss contribution from positive pairs.}
\label{lambda}
\end{table}

\begin{figure}[!t]
\centering
\includegraphics[width=0.5\textwidth]{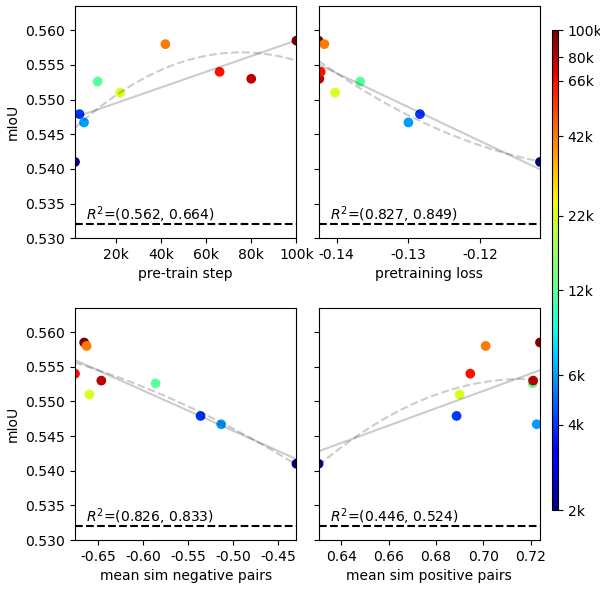}
\caption{Model {\it mIoU} w.r.t. pre-training performance. For each presented metric we fit degree 1 (grey) and degree 2 (grey dashed) polynomials and report $R^2$ values.  Black dashed line $mIoU=0.532$ shows the no pre-training baseline.}
\label{fig:pretrain_perf_ablation}
\end{figure}

\begin{table}[!t]
\begin{center}
\begin{tabular}{|l|c|c|c|}
\hline
\diagbox{$w_r$}{$w_d$} & 3 & 5 & 7 \\
\hline \hline
1   & \textbf{0.567} (3)& 0.553 (5) & 0.550 (7) \\
2   & 0.552 (5) & 0.541 (9) & 0.536 (13) \\
\hline
\end{tabular}
\end{center}
\caption{Model {\it mIoU} for different CSCL window parameters and corresponding $w_d^{eff}$ shown in parenthesis.}
\label{window_params}
\end{table}

Finally, we perform three ablations on CSCL hyperparameters using our dataset in France for 2018. We find that the method is robust to varying parameters as we improve upon the no pre-training baseline ({\it mIoU}=$0.532$) even when initializing with a model pre-trained for only 1 epoch ({\it mIoU}=$0.541$) which is the least amount of pre-training tested. Overall, our ablation study highlights the significance of negative pairs in achieving good performance.

In Table \ref{lambda} we compare results {\bf varying the weight of loss from positive pairs $\lambda$}. We find smaller weights generally leading to a better performance than larger ones. However, best performance is achieved by $\lambda=0.125$ which is close to the ratio of negative to positive pairs in our training set ($0.111$).

The {\bf relationship between pre-training performance and final segmentation performance} is examined in Fig.\ref{fig:pretrain_perf_ablation}. This is done by sampling initializations from various pre-taining steps. For each presented metric we fit degree 1 and 2 polynomials, plot the regression line and report the coefficient of determination $R^2$ indicating the proportion of variance in mIoU that can be predicted by each pre-train metric following these simple regression models. As expected, segmentation performance is generally found to improve with more pre-train steps. We further demonstrate that lower loss and more heavily contrasted negative pairs during pre-training lead to a better final performance and are better predictors of mIoU than pre-training steps. There appears to be no relationship between the mean similarity for positive pairs and final performance. 

Results for {\bf varying CSCL window parameters $w_d$, $w_r$} are presented in Table \ref{window_params}. We find a clear pattern indicating that windows covering smaller effective receptive fields ($w_{d}^{eff} = (w_d -1) * w_r + 1$), thus containing harder negative pairs on average, perform better overall. 

In addition to the above ablations we have tested the possibility of \textbf{using CSCL as a secondary loss term} instead of a pretext task for pre-training. We found that this approach does not work well in practice. Models trained with both the CE and CSCL criteria reached a final segmentation {\it mIoU} of no more than $30\%$ in experiments in France. This is significantly less than the baselines and all other ablations tested and suggests that the two loss terms are incompatible.

\subsection{Semantic segmentation at super-resolution}\label{super_res}
In Table \ref{super_res_cscl} we present results for the baseline and CSCL pre-trained models using the $\times 4$ ground truths. There, the "train res." column indicates which data were used both for pre-training and semantic segmentation training, e.g. the same dataset ($\times 1$ or $\times 4$) is always used for both tasks. To show the benefit over training with $\times 4$ ground truths and no pre-training we upscale the output of the best pre-trained model from Table \ref{satellite_results_main} ($\times 1$) via nearest neighbour interpolation. We note that $\times 4$ with pre-training outperforms the upscaled model ({\it mIoU} 0.597 vs 0.586) and that pre-training has a more significant effect that training at $\times 4$ resolution alone with no pre-training ({\it mIoU} 0.586 vs 0.581). 

\begin{figure}[!t]
    \centering
    \includegraphics[width=0.95\linewidth]{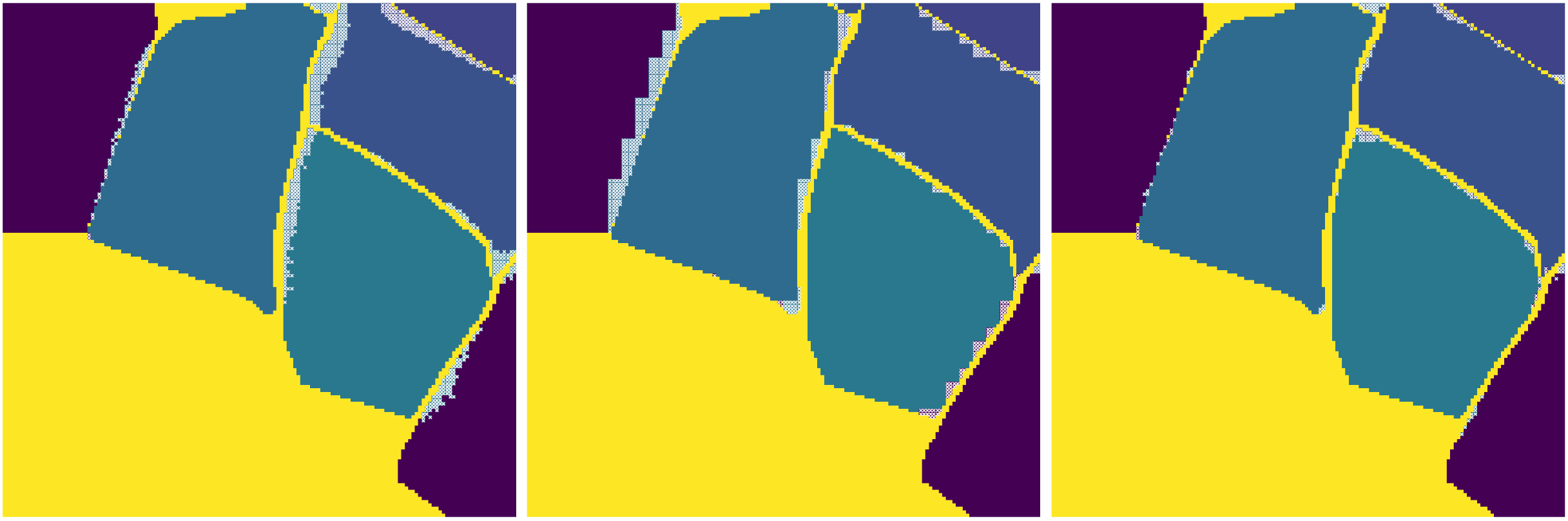}%
    \caption{Qualitative comparison of super-resolution models from Table \ref{super_res_cscl}. (left to right) UNet3Df-$\times 4$, UNet3Df-CSCL-$\times 1$, UNet3Df-CSCL-$\times 4$. White "x" indicates false prediction.}
    \label{qualitative_model_comparison_superres}
\end{figure}

Fig.\ref{qualitative_model_comparison_superres} shows a qualitative comparison of predictions using the models from Table \ref{super_res_cscl}. We note the clustering of false predictions in blocks for the non pre-trained upscaled output and the performance improvement at boundaries for the CSCL vs the randomly initialized model.

\begin{table}[!t]
\begin{center}
\begin{tabular}{|c|c|c|c|c|c|c|c|c|}
\hline
model & train res. & Acc. & mIoU & F1\\
\hline \hline
UNET3Df & $\times 4$ & $66.2 \pm 0.5$ & $58.1 \pm 0.2$ & $69.1 \pm 0.2$ \\
UNET3Df-CSCL & $\times 1$ & $66.9 \pm 0.2$  &  $58.6 \pm 0.2$  &  $69.8 \pm 0.2$  \\
UNET3Df-CSCL & $\times 4$ & \textbf{$67.7 \pm 0.4$}  &  \textbf{$59.7 \pm 0.3$}  &  \textbf{$70.9 \pm 0.2$}  \\
\hline
\end{tabular}
\end{center}
\caption{Super-resolution model performance. The output of models trained at base ($\times 1$) resolution is upscaled to match $\times 4$ ground truths via nearest neighbour interpolation. Presented metrics are mean and $95\%$ confidence intervals among 3 runs. }
\label{super_res_cscl}
\vspace{-0.2cm}
\end{table}

\section{Conclusion}
In this study, we tackled the problem of crop type semantic segmentation from satellite images. Motivated by an identified performance drop at object boundaries we proposed CSCL, a pre-training scheme targeting the ability of a CNN to model the relationship between each input location and its local neighborhood. Using our method we improved the state-of-the-art performance in crop type semantic segmentation. 

Finally, we create a very large dataset of SITS for crop type semantic segmentation which also enables training of semantic segmentation models at higher resolution than the highest resolution S2 band. We make this data publicly available to facilitate further research into this domain.

{\small
\bibliographystyle{IEEEtran}
\bibliography{ms}
}

\end{document}